\documentclass[10pt,twocolumn,letterpaper]{article}

\usepackage[pagenumbers]{cvpr} 

\definecolor{cvprblue}{rgb}{0.21,0.49,0.74}
\usepackage[pagebackref,breaklinks,colorlinks,allcolors=cvprblue]{hyperref}

\usepackage{graphicx}
\usepackage{subcaption}
\usepackage{amsmath}
\usepackage{multirow}
\usepackage{makecell}

\def\paperID{6985} 
\def\confName{CVPR}
\def\confYear{2026}

\title{SeaCache: Spectral-Evolution-Aware Cache for Accelerating Diffusion Models}

\newcommand\blfootnote[1]{%
  \begingroup
  \renewcommand\thefootnote{}\footnote{#1}%
  \addtocounter{footnote}{-1}%
  \endgroup
}

\author{
Jiwoo Chung\textsuperscript{1,\dag}\quad
Sangeek Hyun\textsuperscript{1}\quad
MinKyu Lee\textsuperscript{1}\quad
Byeongju Han\textsuperscript{2}\\
Geonho Cha\textsuperscript{2}\quad
Dongyoon Wee\textsuperscript{2}\quad
Youngjun Hong\textsuperscript{2,*}\quad
Jae-Pil Heo\textsuperscript{1,*}\\[0.2cm]
\textsuperscript{1}Sungkyunkwan University\quad
\textsuperscript{2}NAVER Cloud
}

\begin{document}
\maketitle
\blfootnote{\textsuperscript{\dag} This work was done during an internship at NAVER Cloud.}
\blfootnote{\textsuperscript{*} Co-corresponding authors.}

\begin{abstract}
Diffusion models are a strong backbone for visual generation, but their inherently sequential denoising process leads to slow inference. Previous methods accelerate sampling by caching and reusing intermediate outputs based on feature distances between adjacent timesteps. However, existing caching strategies typically rely on raw feature differences that entangle content and noise. This design overlooks spectral evolution, where low-frequency structure appears early and high-frequency detail is refined later.
We introduce Spectral-Evolution-Aware Cache (SeaCache), a training-free cache schedule that bases reuse decisions on a spectrally aligned representation. Through theoretical and empirical analysis, we derive a Spectral-Evolution-Aware (SEA) filter that preserves content-relevant components while suppressing noise.
Employing SEA-filtered input features to estimate redundancy leads to dynamic schedules that adapt to content while respecting the spectral priors underlying the diffusion model. Extensive experiments on diverse visual generative models and the baselines show that SeaCache achieves state-of-the-art latency-quality trade-offs.
Codes are available at \href{https://github.com/jiwoogit/SeaCache}{github.com/jiwoogit/SeaCache}.
\end{abstract}
    
\section{Introduction}
\label{sec:intro}

Recent diffusion~\cite{songdenoising,sohl2015deep,dhariwal2021diffusion,song2019generative,rombach2022high} and rectified-flow (RF)~\cite{esser2024scaling,liu2023flow} models produce high-quality images and videos through iterative denoising.
Despite this progress, sampling still requires tens to hundreds of steps, which turns user-facing applications into latency bound.
A common remedy is to reduce the step count or the per-step cost through distillation~\cite{meng2023distillation,salimansprogressive,kim2025autoregressive,sauer2024adversarial,sauer2024fast,ding2025efficient}, quantization~\cite{zhangsageattention,chen2025q, shang2023post,yang20241}, or efficient attention~\cite{xisparse,yang2025sparse,zhang2025training,yuan2024ditfastattn,xia2025training}. These approaches are effective but introduce added training overhead and dependence on task or data-specific tuning.

A complementary direction exploits redundancy between consecutive steps via caching.
Caching reduces the number of forward passes by reusing intermediate features from previous timesteps.
Early work adopts static schedules~\cite{zhaoreal,liu2025reusing,li2023faster} that cache features at fixed intervals along the trajectory, which yields predictable speedups.
More recent methods introduce dynamic schedules~\cite{liu2025timestep,adnan2025foresight,bu2025dicache} that decide when to reuse based on the distance between current and cached features, thereby reducing the error introduced by caching.
These approaches focus on where to cache, for example which layers or blocks, while the error itself is still measured in the raw feature space.

\begin{figure}[t]
  \centering
  \begin{subfigure}[t]{1\linewidth}
    \centering
    \includegraphics[width=1\linewidth, trim=0mm 25mm 6mm 0mm, clip]{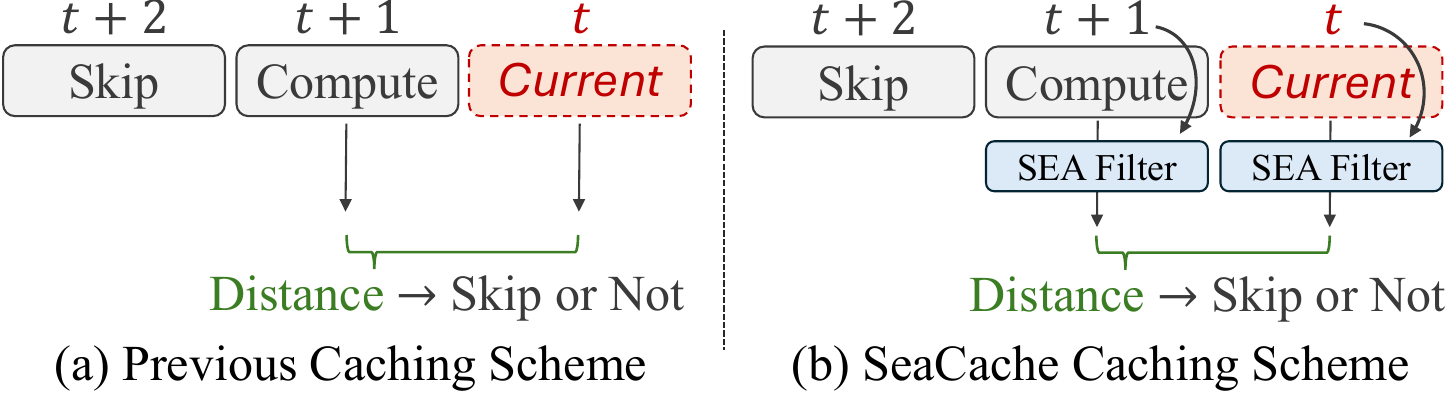}
    \label{fig:intro}
  \end{subfigure}

  \vspace{-8pt}
  \begin{subfigure}[t]{1\linewidth}
    \centering
    \includegraphics[width=1\linewidth]{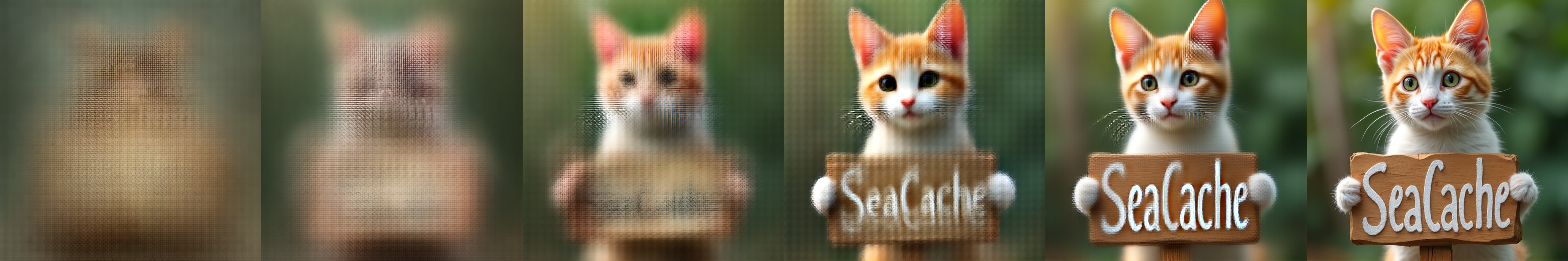}
    \label{fig:x0s}
  \end{subfigure}
  \vspace{-15pt}
  \caption{
  Conceptual illustration and motivation of the proposed caching scheme~(SeaCache) compared with previous caching schemes.
  The lower panel shows a denoising trajectory of a cat image where coarse low-frequency structure appears at early steps and fine high-frequency details emerge at later steps, illustrating the spectral evolution of iterative generative models.
  SeaCache applies a Spectral-Evolution-Aware~(SEA) Filter to raw diffusion features so that the distance measure better captures timestep-aware spectral residuals between timesteps.
  }
  \label{fig:intuition_figure}
  \vspace{-5pt}
\end{figure}

However, these approaches measure errors directly in the raw feature space and overlook \emph{spectral evolution}, a key prior underlying the denoising process.
Independent of caching, prior studies~\cite{lee2025beta,huang2024blue,falck2025fourier,yu2025dmfft} have provided clear evidence that diffusion models exhibit spectral evolution, where ujkearly timesteps establish low-frequency structure and later timesteps refine high-frequency detail, as also illustrated in the lower panel of Fig.~\ref{fig:intuition_figure}.
From this viewpoint, spectral evolution at a given timestep can be interpreted as a change in the signal-to-noise ratio. 
We use the term \emph{signal} for the content-carrying component that is aligned with the clean sample and mainly lies in low frequencies, and \emph{noise} for the residual component that is concentrated in high frequencies and reflects stochastic variation.

In this paper, we incorporate this spectral evolution, or equivalently the evolution of the signal-to-noise ratio, into cache scheduling. 
Rather than treating all spectral components equally, we design a cache metric that focuses on the signal component while downweighting the noise component. 
By grounding reuse decisions on discrepancies in the synthesized content, the resulting metric becomes less sensitive to high-frequency noise and encourages cache gating to respond to meaningful signal alignment rather than stochastic variation.

\begin{figure}[t]
  \centering
  \begin{subfigure}[t]{0.98\linewidth}
    \centering
    \includegraphics[width=0.98\linewidth]{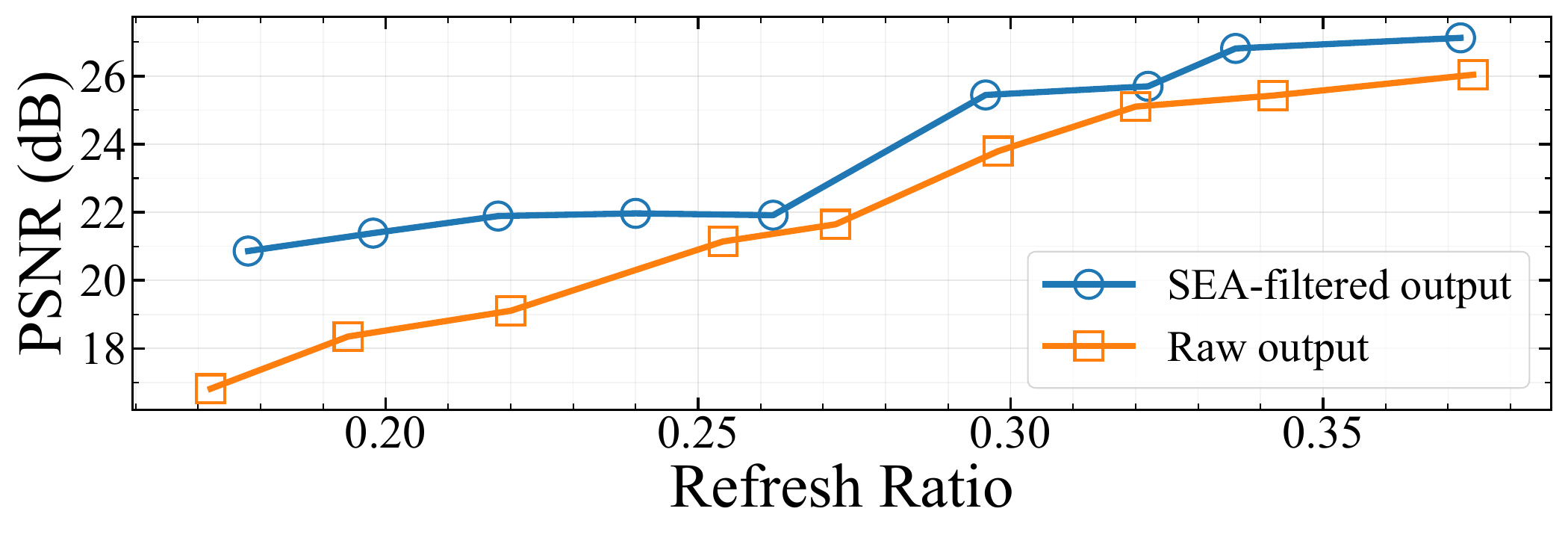}
    \vspace{-3pt}
    \caption{Latency-quality trade-off on \textit{FLUX}.}
    \label{fig:oracle_flux}
  \end{subfigure}
  \begin{subfigure}[t]{0.98\linewidth}
    \centering
    \includegraphics[width=0.98\linewidth]{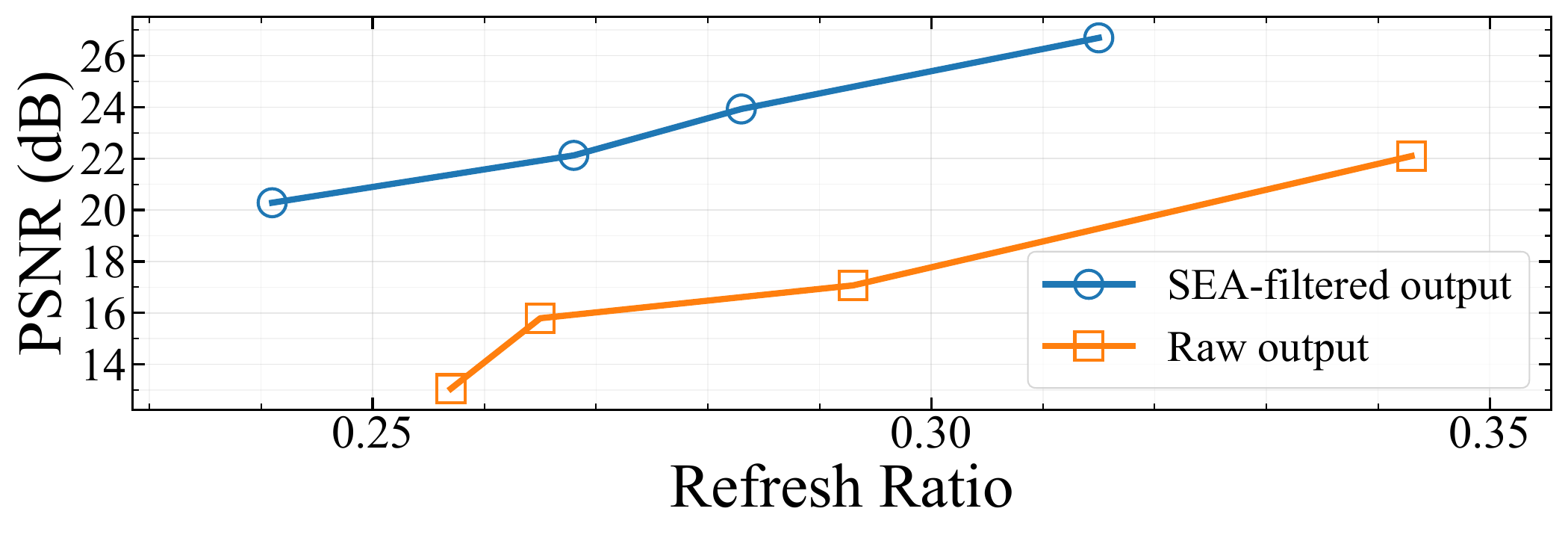}
    \vspace{-3pt}
    \caption{Latency-quality trade-off on \textit{Wan2.1 1.3B}.}
    \label{fig:oracle_wan}
  \end{subfigure}
    \vspace{-4pt}
  \caption{
  \textbf{Latency-quality trade-off in oracle experiments.}
  We compare cache decisions based on raw output differences and SEA-filtered output differences (Sec.~\ref{sub:lpf}) on \textit{FLUX}~\cite{flux2024,labs2025flux1kontextflowmatching} and \textit{Wan2.1 1.3B}~\cite{wan2025}. 
    The refresh ratio is the fraction of timesteps that run a full denoiser evaluation instead of reusing cached features. For each criterion, PSNR is computed between the cached sample and the corresponding full timestep (no-cache) sample, averaged over each prompt set~\cite{saharia2022photorealistic,huang2024vbench}.
  At matched refresh ratios, the filtered criterion consistently achieves higher PSNR with respect to the full-compute trajectory, validating the effectiveness of a spectrum-aware distance for cache scheduling.
  }
  \label{fig:oracle}
\end{figure}

To validate this idea, we conduct an oracle experiment that compares cache schedules derived from raw feature distances with those derived from distances in a signal-emphasized space. 
In standard caching schemes, the decision to skip or compute is based on the distance between input features at consecutive timesteps. 
In our oracle analysis, we instead compare consecutive \emph{output features}, thereby removing input-to-output approximation error and isolating the effect of spectral filtering.
Specifically, we compare two criteria: one th at measures distances after applying the SEA (Spectral-Evolution-Aware) filter, which downweights the noise component (Sec.~\ref{sub:lpf}), and another that uses unfiltered raw outputs, as shown in Fig.~\ref{fig:oracle}.
The filtered criterion yields cache decisions that more closely track the full-compute trajectory, as evidenced by consistently higher PSNR. This suggests that spectrum-aware scheduling better preserves the behavior of the original model.

To this end, we propose Spectral-Evolution-Aware Cache (\textbf{SeaCache}), a simple yet effective caching scheme that encodes the spectral prior of iterative denoising models through a Spectral-Evolution-Aware (SEA) filter, as illustrated in Fig.~\ref{fig:intuition_figure}. 
The SEA filter provides a practical scheduling policy by allowing cache decisions to be driven by the signal component. 
Before measuring feature distances, SeaCache passes intermediate features through a theoretically motivated, timestep-dependent filter that modulates the frequency response along the sampling trajectory. This operation acts as a lightweight reweighting that amplifies the content-relevant signal while downweighting noise-dominated components.

SeaCache is plug-and-play: it requires no architectural modification or retraining, and can be attached to existing caching policies by inserting a single filtering step before distance computation. 
The method is both network-agnostic and sampler-agnostic, enabling integration across diverse diffusion and rectified-flow models. 
In practice, SeaCache substantially reduces the number of forward passes while preserving the perceptual fidelity of the original outputs, and it consistently improves the latency-quality trade-off over prior caching schemes across experiments.

Our main contributions are threefold.
\begin{itemize}
    \item We propose \textbf{SeaCache}, a simple yet effective caching policy that bases reuse decisions on a timestep-aligned spectral representation of the generative trajectory.
    \item We revisit prior caching strategies and show that raw feature metrics ignore spectral evolution, while our formulation bases cache decisions on content rather than noise.
    \item Extensive experiments on multiple visual generative models show that our method achieves better latency–quality trade-offs than prior caching baselines.

\end{itemize}

\section{Related Work}
\label{sec:related}

\subsection{Generative Model Acceleration}
Recent generative models~\cite{songdenoising,sohl2015deep,dhariwal2021diffusion,song2019generative,rombach2022high,esser2024scaling,liu2023flow,chung2025fine} have advanced visual synthesis, but their multi-step denoising procedures make inference latency and computation a primary bottleneck. Step reduction methods compress the sampling trajectory using improved solvers~\cite{songdenoising,lu2022dpm,zhao2023unipc} and distillation-based samplers~\cite{luo2023latent,song2023consistency,salimansprogressive}. These approaches are effective but require additional training and often modify the original model. Another line of work reduces the cost of each step through quantization~\cite{shang2023post,he2023ptqd,li2023q,so2023temporal}, efficient attention~\cite{xisparse,yang2025sparse,zhang2025training,dao2022flashattention,dao2023flashattention2,pu2024efficient,yuan2024ditfastattn,xia2025training,anagnostidis2025flexidit}, and token reduction~\cite{so2023temporal,kim2024token,zhang2025training}. These techniques lower FLOPs while preserving the sequential dependency of the sampler, but they typically demand extra resources and engineering effort. This limitation motivates caching-based acceleration, which exploits redundancy across successive timesteps to reuse intermediate features without additional training.

\subsection{Caching-based Acceleration}
Caching-based acceleration reuses intermediate computations across adjacent timesteps without retraining. Early methods~\cite{ma2024deepcache,li2023faster,wimbauer2024cache} achieve speedups by reusing features but are designed for U-Net architectures, which limits their applicability to transformer-based models. To address this limitation, later work~\cite{chen2024delta,selvaraju2024fora,liu2025faster} adapts caching to DiT architectures~\cite{selvaraju2024fora,li2023faster} for image synthesis. For video, PAB~\cite{zhaoreal} selects different timestep intervals for each attention block and achieves speedups.

These methods rely on static schedules and cannot adapt to input diversity, so recent work adopts dynamic policies that respond to the generated signal~\cite{liu2025speca,lvfastercache,kahatapitiya2025adaptive,liu2025timestep,ma2025magcache,aggarwal2025evolutionary}. For example, AdaCache~\cite{kahatapitiya2025adaptive} accounts for motion complexity for accelerating video generation. TeaCache~\cite{liu2025timestep} and DiCache~\cite{bu2025dicache} estimate output changes from distances measured near the input features and assume that these distances provide a reliable redundancy signal between adjacent-timesteps.
In our work, we measure redundancy in a timestep-aligned spectral space that emphasizes content-carrying components.
Unlike prior dynamic caching, SeaCache explicitly models \emph{spectral evolution} through a timestep-conditioned SEA filter motivated by a linear-denoiser view, and applies gain normalization to enable stable distance measurements across timesteps.
As a result, SeaCache is the first caching policy that injects an explicit frequency prior into the reuse decision.

Recent studies~\cite{zou5584552feb,liu2025freqca,liu2025speca} explore reusing features differently across frequency bands. In contrast, we focus on when to reuse rather than how to utilize cached features.
Leveraging the spectral evolution prior where low-frequency structure emerges early while high-frequency details are refined later, we propose a simple cache policy that plugs easily into existing caching baselines.

\section{Preliminary}
\label{sec:prelim}

\subsection{Denoising Generative Models}
Diffusion probabilistic models (DPMs)~\cite{ho2020denoising} and rectified flow (RF) models~\cite{liu2023flow} generate samples by iteratively removing noise.
Let \(X\) denote a clean image or video, and let an encoder map \(X\) to a latent \(x_0\).
For images, we denote \(x_0 \in \mathbb{R}^{H \times W \times C}\), and for videos \(x_0 \in \mathbb{R}^{H \times W \times F \times C}\), where \(H, W, F,\) and \(C\) denote the height, width, number of frames, and channels of the latent representation, respectively.

We adopt the standard forward noising model at discrete solver steps \(t \in \{0,\ldots,T\}\):
\begin{equation}
\label{eq:lin}
x_t \;=\; a_t x_0 + b_t \varepsilon,\qquad \varepsilon \sim \mathcal{N}(0,\mathbf{I}),
\end{equation}
where \(T\) is the total number of steps and \((a_t,b_t)\) are determined by the noise schedule.
For DPMs~\cite{ho2020denoising}, \(a_t = \sqrt{\bar\alpha_t}\) and \(b_t = \sqrt{1-\bar\alpha_t}\) with \(\bar\alpha_t \in [0,1]\) given by the schedule.
For RFs~\cite{liu2023flow}, the same linear mixture provides a useful approximation with \(a_t = 1-\alpha_t\) and \(b_t = \alpha_t\), where \(\alpha_t = \tfrac{t}{T}\).

Under this noise mixture model, DPMs are trained to predict the noise \(\varepsilon\) from the noised latent \(x_t\) at timestep \(t\).
The corresponding training objective is
\begin{equation}
    \mathcal{L}_\text{DPM}
    = \mathbb{E}_{x_0,\,t,\,\varepsilon,\,y}
      \big[\big\|\varepsilon - \epsilon_\theta(x_t, t, y)\big\|_2^2\big],
\end{equation}
where \(y\) is a conditioning signal and \(\epsilon_\theta\) is a denoising network that estimates the noise added to \(x_0\).
Sampling proceeds in reverse, starting from \(x_T \approx \varepsilon\) and iteratively reconstructing \(x_0\).
This iterative denoising process induces strong redundancy between outputs at adjacent timesteps, and cache-based acceleration exploits this redundancy by reusing intermediate predictions.

\subsection{Timestep-Aware Dynamic Caching}
\label{sub:teacache}
A recent approach, TeaCache~\cite{liu2025timestep}, quantifies change at step \(t\) using the timestep-modulated input \(I_t = \phi(x_t, t)\), where \(\phi\) injects a timestep embedding into the input \(x_t\). This proxy is strongly correlated with the denoiser output \(O_t\) while remaining inexpensive to compute, and for brevity we refer to \(I_t\) as the input feature. The relative \(\ell_1\) distance is then defined as
\begin{equation}
\label{eq:rel_l1}
\Delta_t = \mathrm{L1}_{\mathrm{rel}}(I_t, I_{t+1})
= \frac{\lVert I_t - I_{t+1} \rVert_1}{\lVert I_{t+1} \rVert_1 + \xi},
\end{equation}
with a small constant \(\xi\) for numerical stability~\cite{ma2024deepcache,liu2025timestep,bu2025dicache}.

After computing the model output at step \(t_a\), the same output is reused for steps \(t \in [t_a, t_b - 1]\) until the accumulated change exceeds a threshold~$\delta$.
Let \(t_b > t_a\) be the smallest index that satisfies
\begin{equation}
\label{eq:accum-th}
\sum_{s=t_a}^{t_b-1} \Delta_s \le \delta < \sum_{s=t_a}^{t_b} \Delta_s,
\end{equation}
at which point a refresh is triggered at \(t_b\) and the accumulator is reset.
Smaller \(\delta\) leads to more frequent refreshes and higher fidelity, while larger \(\delta\) increases speed at the risk of artifacts.
We follow the accumulated-distance rule and keep the same refresh logic on the timestep-modulated feature at the pre-attention input of the first transformer block to maximize skipped computation, as in TeaCache.
Accordingly, SeaCache injects spectral priors into \(\Delta_t\) by measuring change in a frequency-aware filtered representation.

\begin{figure*}[t!]
  \centering
  \includegraphics[width=1.0\textwidth, trim=0mm 40mm 38mm 0mm, clip]{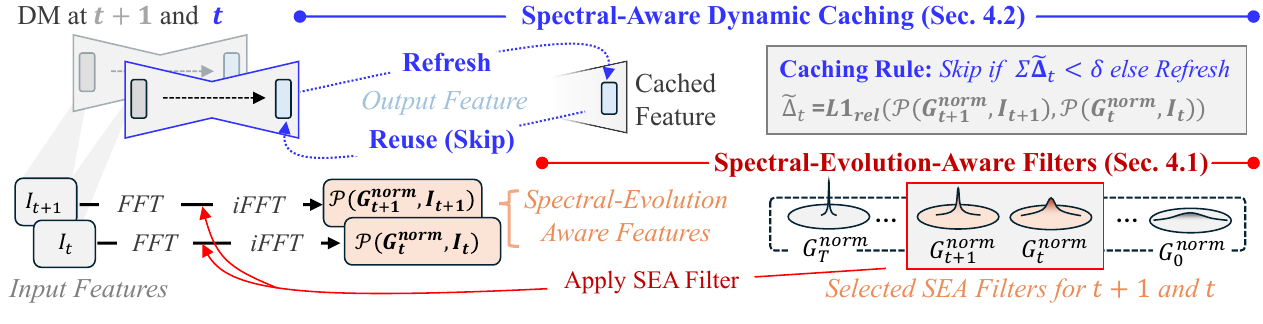}
  \vspace{-20pt}
  \caption{
  \textbf{Overview of SeaCache.}
  Given input features \(I_t\) and \(I_{t+1}\), SeaCache first applies FFT, multiplies by the timestep-dependent SEA filters \(G_t^{\mathrm{norm}}\) and \(G_{t+1}^{\mathrm{norm}}\), and then applies iFFT to obtain spectral-evolution-aware features \(\mathcal{P}(G_t^{\mathrm{norm}}, I_t)\) and \(\mathcal{P}(G_{t+1}^{\mathrm{norm}}, I_{t+1})\) (Sec.~\ref{sub:lpf}). 
  A spectrum-aware dynamic caching module (Sec.~\ref{sub:schedule}) measures the relative distance \(\widetilde{\Delta}_t\) between consecutive filtered features, accumulates it over timesteps, and either reuses the cached output or refreshes the denoiser when the threshold \(\delta\) is exceeded. 
  The underlying diffusion model remains unchanged, so SeaCache acts as a plug-and-play cache policy that replaces only the distance metric.
  }
  \label{fig:main_figure}
  \vspace{-5pt}
\end{figure*}

\section{Method: SeaCache}
\label{sec:method}
Prior analyses~\cite{lee2025beta,huang2024blue,anagnostidis2025flexidit} and the lower panel of Fig.~\ref{fig:intuition_figure} reveal a form of spectral evolution in diffusion models, where early steps build low-frequency structure and later steps refine high-frequency detail.
Motivated by this behavior, we design a spectrum-aware reuse metric that guides cache scheduling across timesteps.
Our approach proceeds in three stages (Fig.~\ref{fig:main_figure}).
First, in Sec.~\ref{sub:lpf}, we formalize the denoiser frequency response and derive a timestep-dependent filter that captures this evolution.
Second, in Sec.~\ref{sub:schedule}, we introduce an input proxy whose filtered distance is closely related to the filtered output distance, which enables a training-free, plug-and-play schedule.
Finally, we replace the original metric \(\Delta_t\) with its spectrum-aware counterpart \(\widetilde{\Delta}_t\) while preserving the standard accumulated distance based refresh rule.

\begin{figure}[t]
  \centering
  \begin{subfigure}[t]{0.98\linewidth}
    \centering
    \includegraphics[width=0.98\linewidth]{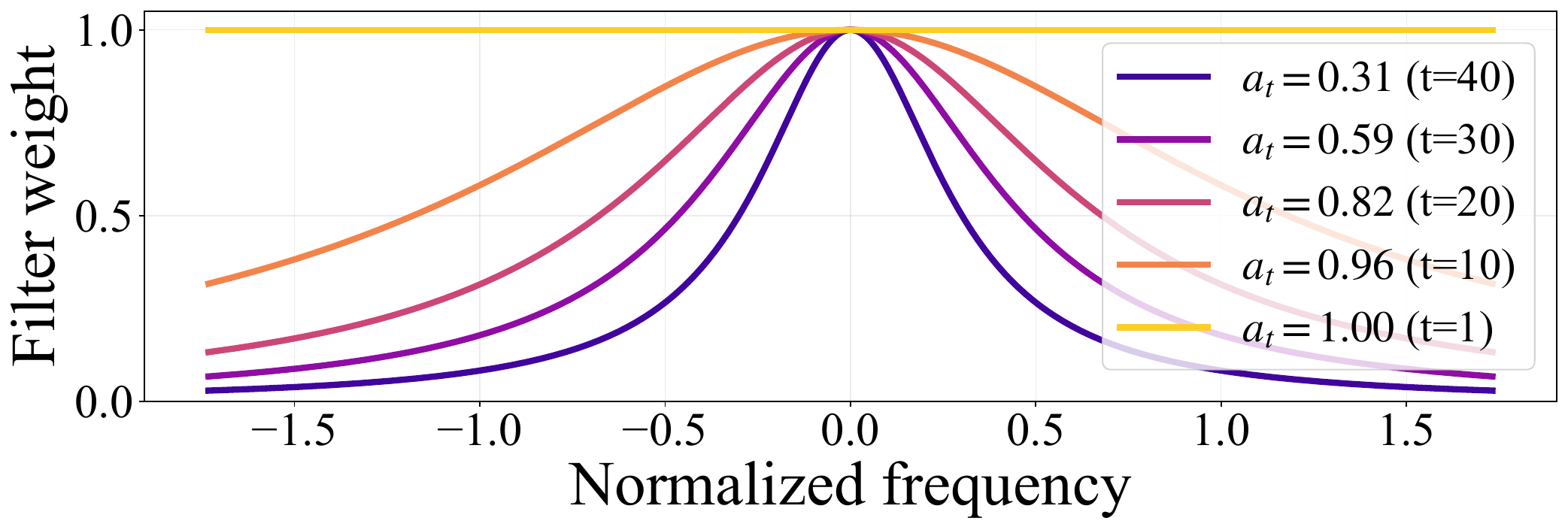}
    \vspace{-2pt}
    \caption{Optimal linear denoising filter \(G_t(f)\) for different \(t\).}
    \label{fig:lpf_rf_peak}
  \end{subfigure}
  \begin{subfigure}[t]{0.98\linewidth}
    \centering
    \includegraphics[width=0.98\linewidth]{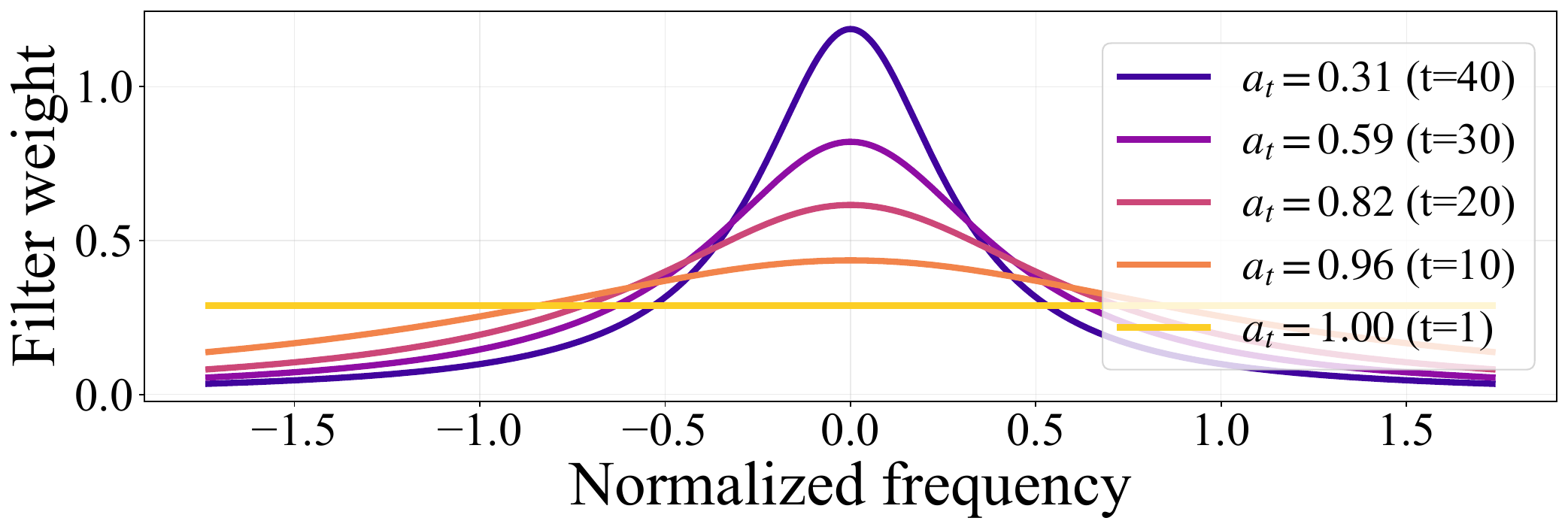}
    \vspace{-2pt}
    \caption{Normalized SEA filter \(G^{\mathrm{norm}}_t(f)\) for different \(t\).}
    \label{fig:lpf_rf_density}
  \end{subfigure}
  \vspace{-4pt}
  \caption{
  \textbf{Visualization of timestep-dependent denoising filters.}
  (a) Optimal linear denoising responses \(G_t(f)\) across timesteps, where early steps primarily pass low-frequencies and later steps gradually include higher frequencies, reflecting spectral evolution. 
  (b) Corresponding normalized filters \(G^{\mathrm{norm}}_t(f)\) with unit mean gain, which stabilize filtered feature energy across timesteps and are used as SEA filters for cache scheduling.
  }
  \vspace{-3pt}
  \label{fig:lpf_vis}
\end{figure}

\subsection{Spectral-Evolution-Aware Filter}
\label{sub:lpf}
To design a filter that reflects spectral evolution, we formalize how the effective frequency band changes across timesteps. Motivated by Spectral Diffusion~\cite{yang2023diffusion}, we adopt the timestep-dependent frequency response derived under the optimal linear denoiser. For notational simplicity, we describe a single-channel 2D filter. In implementation, the filter is applied per-channel over the spatial (2D) axes for images and over the spatiotemporal (3D) axes for videos.

We consider the linear minimum mean squared error (MMSE) estimator \(\widehat{x}_0 = h_t \ast x_t\) obtained by minimizing
\(
J_t(h_t) = \|h_t \ast x_t - x_0\|_2^2,
\)
where \(h_t\) is a linear denoising filter and \(\ast\) denotes convolution (which corresponds to pointwise multiplication in the frequency domain).
We denote by \(h_t^\star\) the optimal linear filter that minimizes \(J_t\). Let \(G_t(f)\) denote the frequency response of the optimal linear denoising filter \(h_t^{\star}\) at frequency \(f\), and let \(S_x(f)\) denote the power spectrum of \(x_0\) at frequency \(f\). Under the linear mixture in Eq.~\ref{eq:lin}, the optimal frequency response of \(h_t^{\star}\) takes a Wiener-like form~\cite{wiener1964extrapolation}:
\begin{equation}
\label{eq:wiener-general}
G_t(f) = \frac{a_t\,S_x(f)}{a_t^2\,S_x(f) + b_t^2},
\end{equation}
and although the linearity assumption on \(h_t\) is restrictive, it still provides useful insight into spectral evolution.

Assuming a natural power law spectrum~\cite{burton1987color,field1987relations,tolhurst1992amplitude,van1996modelling} for \(S_x(f)\), representative DPM responses \(G_t(f)\) are shown in Fig.~\ref{fig:lpf_rf_peak}.
In the reverse diffusion process, \(t\) decreases from \(T\) to \(0\) while \(a_t\) increases from \(0\) to \(1\), gradually recovering high frequency detail in a way that is consistent with spectral evolution.
The resulting filters for DPMs and RF models exhibit nearly identical behavior, and for brevity we present the analysis in terms of the DPM in the main text.
Full derivations are provided in the supplementary material. In this view, we refer to the low-frequency content-carrying component aligned with the clean sample as the signal and to the high-frequency residual that primarily reflects stochastic variation as noise.

We formulate the optimal response \(G_t(f)\) and confirm spectral evolution under this model. 
We then use \(G_t(f)\) to filter features in the frequency domain for constructing a spectrum-aware representation that emphasizes the signal component while suppressing noise.
Specifically, we define a feature-level mapping \(\mathcal{P}_t\) by applying the fast Fourier transform (FFT~\cite{nussbaumer1981fast}), multiplying by the timestep-dependent spectrum-aware filter \(G_t(f)\), and returning to the original space via the inverse FFT (iFFT):
\begin{equation}
\label{eq:pt-def-1d}
\mathcal{P}(G_t,I_t)
\;=\;\mathrm{iFFT}\big(G_t(f)\ \odot\ \mathrm{FFT}(I_t)\big),
\end{equation}
where \(f\) indexes radial frequencies on the discrete Fourier grid and \(\odot\) denotes element-wise multiplication with broadcasting across channels and spatial or spatiotemporal dimensions. 
This operator \(\mathcal{P}_t(G_t,\cdot)\) induces a timestep-dependent passband and defines the filtered feature space in which spectrum-aware cache distances are computed.

Before filtering the features, the raw response \(G_t\) exhibits a timestep-dependent gain, as in the varying radial averages in Fig.~\ref{fig:lpf_rf_peak}. 
To ensure that distances are comparable across timesteps, we normalize this gain by enforcing a constant mean over radial frequencies, yielding density-normalized response~\(G_t^{\mathrm{norm}}\)~(Fig.~\ref{fig:lpf_rf_density}).
Specifically, let the discrete radial frequencies on the FFT grid be
\(\mathcal{F} = \{f_\ell\}_{\ell=0}^{L-1}\), where \(L\) is the number of radial bins, induced by the spatial resolution \(H \times W\) for images.
We define
\begin{equation}
    \nu_t = \Bigl(\frac{1}{L}\sum_{f_\ell\in\mathcal{F}} G_t(f_\ell)\Bigr)^{-1},
    \quad
    G_t^{\mathrm{norm}}(f) = \nu_t\, G_t(f),
\end{equation}
where \(\nu_t\) is average energy over radial bins such that \(G_t^{\mathrm{norm}}\) has constant mean gain over \(\mathcal{F}\).
Using this density normalized filter, we empirically observe that distances computed after filtering better reflect denoising redundancy than their raw counterparts, and we use \(G_t^{\mathrm{norm}}\) for our SEA filter in the subsequent caching schedule.

\begin{figure}[t]
  \centering
  \begin{subfigure}[t]{0.98\linewidth}
    \centering
    \includegraphics[width=0.98\linewidth]{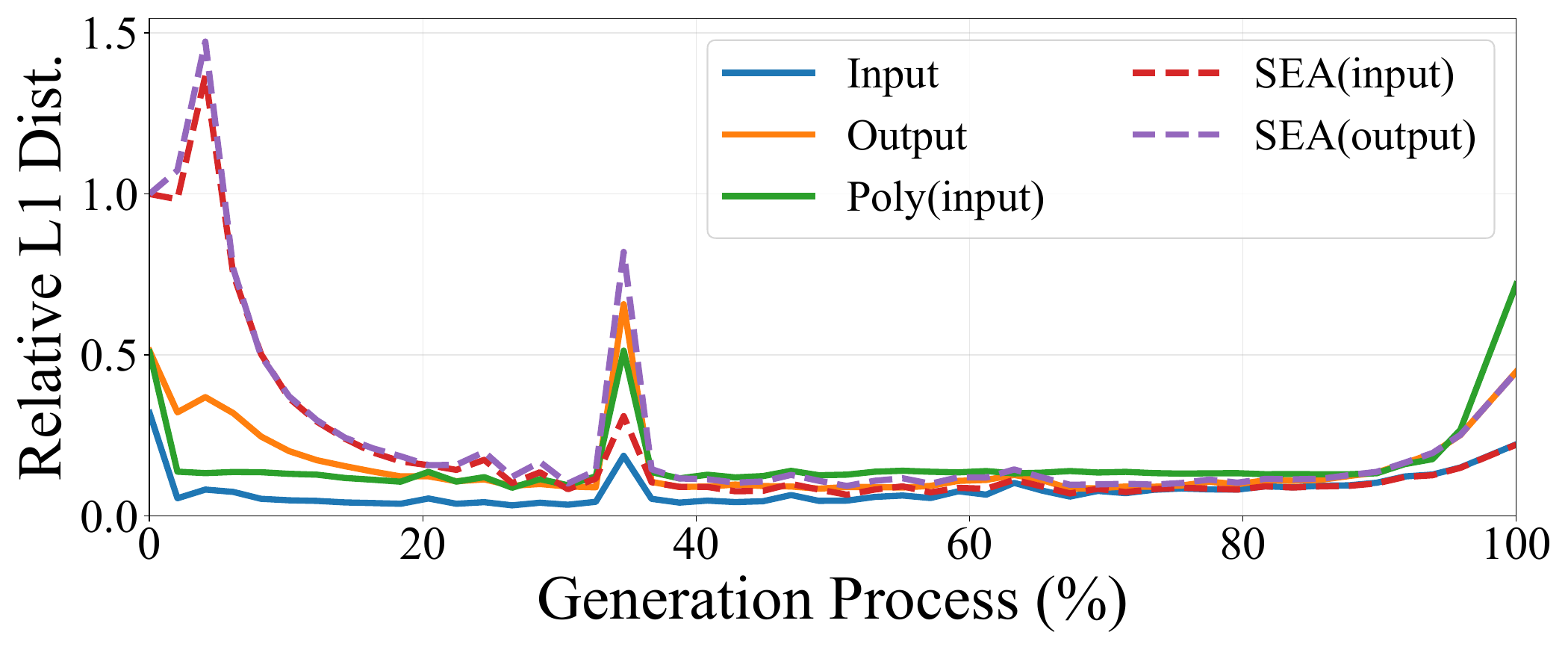}
    \vspace{-2pt}
    \caption{Relative $\ell_1$ across the generation process on \textit{FLUX}.}
    \label{fig:lines_raw}
  \end{subfigure}\hfill
  \begin{subfigure}[t]{0.98\linewidth}
    \centering
    \includegraphics[width=0.98\linewidth]{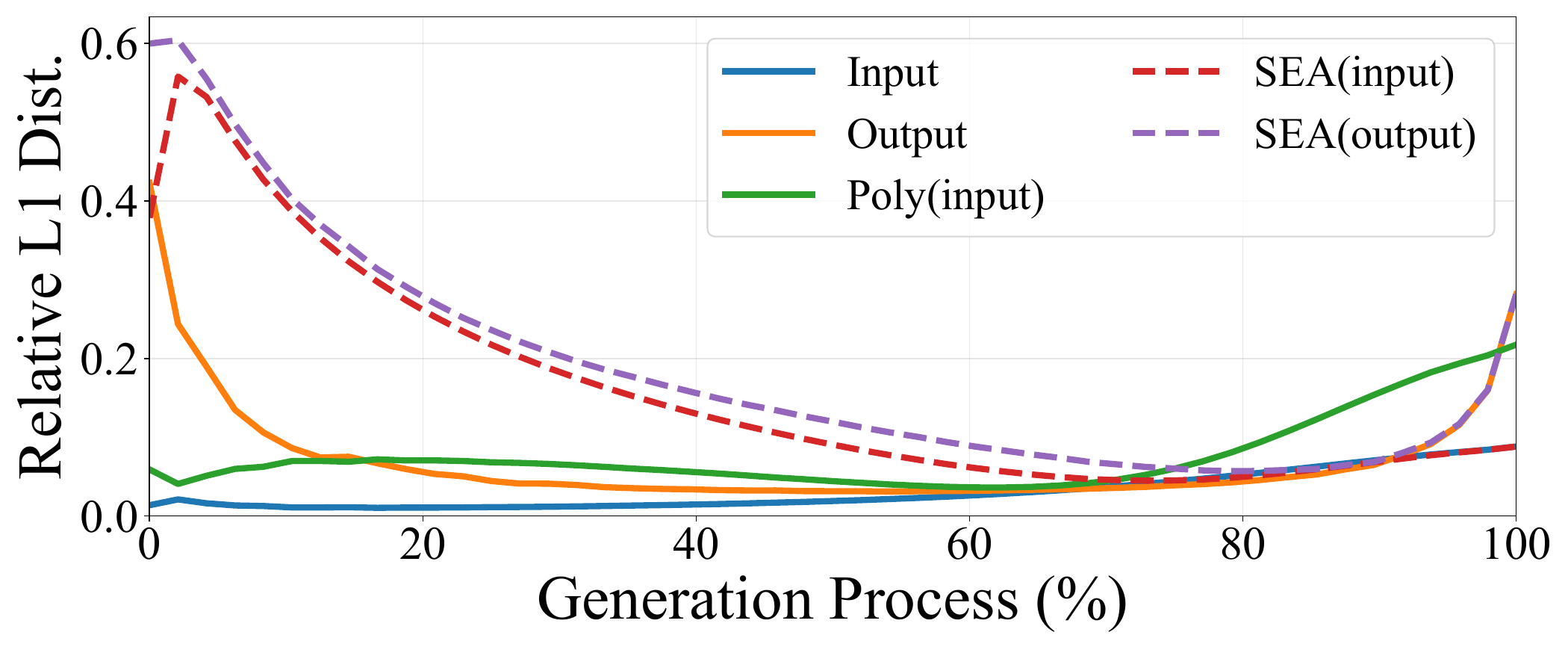}
    \vspace{-2pt}
    \caption{Relative $\ell_1$ across the generation process on \textit{Wan2.1 1.3B}.}
    \label{fig:lines_filtered}
  \end{subfigure}
  \vspace{-5pt}
  \caption{
  \textbf{Relative $\ell_1$ across the generation process.}
  Stepwise relative $\ell_1$ distances between consecutive timesteps for different feature choices, averaged over ten samples for each model.
  \emph{Input} denotes distances on the timestep-modulated input features \(I_t\).
  \emph{Output} is the last block outputs \(O_t\).
  \emph{SEA(Input)}, \emph{SEA(Output)} applies the SEA filter to the input and output features, respectively.
  \emph{Poly(Input)} corresponds to the polynomial-fitted input distance which is designed to approximate output differences from input features.
  SEA-filtered inputs closely track SEA-filtered outputs across timesteps, whereas other inputs show weaker alignment.
  }
  \label{fig:diff_lines}
  \vspace{-7pt}
\end{figure}

\subsection{Spectrum-Aware Dynamic Caching}
\label{sub:schedule}

Prior caching methods~\cite{liu2025reusing,ma2024deepcache,liu2025faster,wimbauer2024cache,bu2025dicache} typically assume that differences between consecutive model outputs reflect redundancy relative to a full-compute trajectory. Building on this assumption, they construct dynamic schedules by approximating output differences from input-side features such as intermediate layers or blocks. However, the oracle study in Sec.~\ref{sec:intro} and Fig.~\ref{fig:oracle} show that this raw feature formulation is suboptimal. Cache decisions based on SEA-filtered outputs stay closer to the full compute trajectory than those based on raw outputs at the same refresh ratio.

Directly using SEA-filtered outputs in the cache metric is not practical, since the output \(O_t\) is only available after a full denoiser run and thus offers no speedup. We therefore seek an input-side proxy that matches the SEA-filtered output distance as closely as possible. Building on the input features \(I_t\), introduced in Sec.~\ref{sub:teacache}, we compare several candidates: raw input \(I_t\), raw output \(O_t\), the polynomial fitted input used in TeaCache~\cite{liu2025timestep}, and their SEA-filtered counterparts obtained by applying \(\mathcal{P}_t(G_t,\cdot)\) from Sec.~\ref{sub:lpf}.

Fig.~\ref{fig:diff_lines} reports the relative \(\ell_1\) distance between consecutive timesteps for these feature choices, averaged over ten samples on \textit{FLUX} and \textit{Wan2.1 1.3B}. The SEA-filtered input distances \(\mathcal{P}_t(G_t, I_t)\) closely follow the SEA-filtered output distances \(\mathcal{P}_t(G_t,O_t)\) along the entire trajectory, while raw input and polynomial fitted input show weaker alignment, especially at early timesteps. Moreover, the SEA-filtered input distances are larger at early timesteps, which is consistent with the common practice of always recomputing early steps in many prior caching schemes~\cite{liu2025reusing,ma2024deepcache,bu2025dicache,ma2025magcache}. This empirical finding is desirable because the SEA filter suppresses stochastic noise while preserving content-carrying components, which makes adjacent-timestep features more stable and faithful proxies for output change.
These results support SEA-filtered inputs as a reliable, training-free proxy for adapting spectrum-aware redundancy.

In SeaCache, distance is therefore measured after density normalized filtering, and the per-step cache metric is defined as
\begin{equation}
\label{eq:rel-l1-lf}
\widetilde{\Delta}_t
= \mathrm{L1}_{\mathrm{rel}}\big(\mathcal{P}(G^\mathrm{norm}_t,I_t),\,\mathcal{P}(G^{\mathrm{norm}}_{t+1},I_{t+1})\big).
\end{equation}
The accumulated distance rule in Eq.~\eqref{eq:accum-th} is kept unchanged. After a refresh at \(t_a\), the cached output is reused for \(t \in [t_a, t_b - 1]\), and the next refresh occurs at the smallest \(t_b\) whose accumulated distance exceeds the threshold \(\delta\). This yields a spectral-evolution-aware, timestep-dependent gate that is training-free and architecture-agnostic, and depends only on the shared sampler schedule coefficients \((a_t, b_t)\).

\section{Experiments}

\begin{table}[t]
  \caption{
  Quantitative comparison in \textit{FLUX.1-dev}~\cite{flux2024,labs2025flux1kontextflowmatching}.
  }
  \vspace{-5pt}
\centering
\setlength{\tabcolsep}{2pt}        
\renewcommand{\arraystretch}{0.95} 
\footnotesize   
\begin{tabular}{lrrrrr}
\toprule
Method & Latency~(s) & TFLOPs & PSNR~\(\uparrow\) & LPIPS~\(\downarrow\) & SSIM~\(\uparrow\) \\
\midrule
Original~(50 steps) & 20.9 & 2976 & -- & -- & -- \\
Vanilla 25 steps & 10.5 & 1487 & 15.553 & 0.409 & 0.668 \\
Vanilla 15 steps & 6.4 & 892 & 17.842 & 0.305 & 0.740 \\
\midrule
TeaCache~(\(\delta\)=0.3) & 11.4 & 1547 & 20.762 & 0.211 & 0.810 \\
TaylorSeer~(\(\mathcal{S}\)=3) & 9.8 & 1191 & 22.783 & 0.163 & 0.828 \\
SeaCache~($\delta$=0.3) & \textbf{9.4} & \textbf{1098} & \textbf{26.285} & \textbf{0.106} & \textbf{0.893} \\\midrule
\(\Delta\)-Dit & 15.5 & 1984 & 17.403 & 0.336 & 0.710 \\
ToCa & 15.9 & 1263 & 18.398 & 0.324 & 0.700 \\
TeaCache~(\(\delta\)=0.6) & 7.1 & 892 & 17.214 & 0.348 & 0.714 \\
TaylorSeer~(\(\mathcal{S}\)=5) & 7.5 & 834 & 19.972 & 0.236 & 0.762 \\
SeaCache~(\(\delta\)=0.6) & \textbf{6.4} & \textbf{773} & \textbf{21.332} & \textbf{0.226} & \textbf{0.798} \\
\bottomrule
\end{tabular}
  \label{tab:main_flux}
\vspace{-4pt}
\end{table}

\begin{table}[t]
\renewcommand{\arraystretch}{0.95} 
  \centering
\small   
  \setlength{\tabcolsep}{3pt}        
    \caption{Comparison of average rank on CycleReward~\cite{bahng2025cycle}.}
      \vspace{-5pt}
  \begin{tabular}{l c | l c}
    \toprule
    Method (\(\approx\)50\%) & Rank~\(\downarrow\) & Method (\(\approx\)30\%) & Rank~\(\downarrow\) \\
    \midrule
    TeaCache~(\(\delta\)=0.3)~\cite{liu2025timestep}   & 2.01 & TeaCache~(\(\delta\)=0.6)~\cite{liu2025timestep}   & 2.07 \\
    TaylorSeer~(\(\mathcal{S}\)=3)~\cite{liu2025reusing} & 2.08 & TaylorSeer~(\(\mathcal{S}\)=5)~\cite{liu2025reusing} & 1.98 \\
    SeaCache~($\delta$=0.3)       & \textbf{1.91} & SeaCache~($\delta$=0.6)       & \textbf{1.96} \\
    \bottomrule
  \end{tabular}
\label{tab:reward_flux}
\vspace{-5pt}
\end{table}

\subsection{Experimental Settings}
\noindent\textbf{Model configurations.}
We evaluate on three state-of-the-art visual generative models. 
FLUX.1-dev~\cite{flux2024,labs2025flux1kontextflowmatching} is a text-to-image model. 
HunyuanVideo~\cite{kong2024hunyuanvideo} and Wan2.1~\cite{wan2025} are text-to-video models. 
For Wan2.1, we use the 1.3B pre-trained checkpoint. 
All models are sampled for 50 steps under their default configurations.
For example, when using TaylorSeer~\cite{liu2025reusing}, we follow its default settings and set the expansion order to 1 for FLUX.1-dev and to 2 for HunyuanVideo and Wan2.1.
FLUX experiments run on NVIDIA Blackwell Pro 6000 GPUs, and HunyuanVideo and Wan2.1 are evaluated on NVIDIA A100 GPUs.

\noindent\textbf{Baseline configurations.}
TeaCache~\cite{liu2025timestep} is applied using the official implementation with default settings, and we adjust the distance threshold \(\delta\) to control the cache ratio.
TaylorSeer~\cite{liu2025reusing} is also used with the official code.
For a fair comparison, we explicitly refresh the first five timesteps for images and the first three timesteps for videos, and we adjust the stride \(\mathcal{S}\) to control the cache ratio.
ToCa~\cite{zouaccelerating} and DiCache~\cite{bu2025dicache} are employed through their official implementations under default settings.
For \(\Delta\)-DiT~\cite{chen2024delta}, we follow the reference implementation provided with TaylorSeer.

\noindent\textbf{Evaluation protocol.}
For all experiments, generated images and videos are stored as PNG and MP4 files, respectively. 
For text-to-image generation, we evaluate 200 DrawBench prompts~\cite{saharia2022photorealistic} and generate \(1024 \times 1024\) images. 
For text-to-video generation, we use 944 prompts from VBench~\cite{huang2024vbench} and generate a 480p video with 65 frames per prompt.
For each configuration, the full timestep output of the original model serves as the reference, and PSNR (computed on RGB values), LPIPS~\cite{zhang2018unreasonable}, and SSIM~\cite{wang2004image} are computed between each cached sample and its reference and then averaged over all samples. 
TFLOPs are measured with Calflops~\cite{calflops} and reported in tera operations. 
We further assess perceptual quality using CycleReward~\cite{bahng2025cycle}, a state-of-the-art image reward benchmark. 
The initial random seed is shared across our method and all baselines, and we consider two cache budgets, approximately \(50\%\) and \(30\%\).

\subsection{Quantitative Comparison}

\noindent\textbf{Text-to-image generation.}
We compare SeaCache with existing caching methods on \textit{FLUX.1-dev}~\cite{flux2024,labs2025flux1kontextflowmatching} in Tab.~\ref{tab:main_flux}. At a moderate budget (roughly \(50\%\) refresh ratio), TeaCache~\cite{liu2025timestep} and TaylorSeer~\cite{liu2025reusing} stay close to the 25 step baseline, while SeaCache further reduces latency and FLOPs and at the same time improves PSNR, LPIPS, and SSIM. This trend persists under stronger acceleration (roughly \(30\%\) refresh ratio). Baselines exhibit clear drops in reconstruction quality, whereas SeaCache achieves the fastest setting among caching methods and still attains the best metrics, yielding a stronger latency-quality trade-off.

We also assess perceptual quality using CycleReward~\cite{bahng2025cycle} in Tab.~\ref{tab:reward_flux}.
At both budgets~(\(\approx 50\%\) and \(\approx 30\%\)), SeaCache achieves the lowest average reward rank among TeaCache and TaylorSeer, showing a better latency-quality trade-off for both reconstruction fidelity and preference.

\begin{table}[t]
\centering
\setlength{\tabcolsep}{2pt}
\renewcommand{\arraystretch}{0.95}
\footnotesize
\caption{Quantitative comparison in \textit{HunyuanVideo}~\cite{kong2024hunyuanvideo}.}
  \vspace{-5pt}
\begin{tabular}{lrrrrr}
\toprule
Method & Latency~(s) & TFLOPs & PSNR~\(\uparrow\) & LPIPS~\(\downarrow\) & SSIM~\(\uparrow\) \\
\midrule
Original (50 steps) & 182.6 & 14038 & -- & -- & -- \\
Vanilla 25 steps & 93.7 & 7019 & 19.97 & 0.263 & 0.731 \\
Vanilla 15 steps & 56.8 & 4211 & 17.49 & 0.371 & 0.662 \\
\midrule
TeaCache~(\(\delta\)=0.12) & 98.5 & 6994 & 23.40 & 0.133 & 0.805 \\
TaylorSeer~(\(\mathcal{S}\)=2) & 96.9 & 7299 & 24.14 & 0.152 & 0.820 \\
SeaCache~(\(\delta\)=0.19) & \textbf{90.8} & \textbf{6747} & \textbf{32.39} & \textbf{0.047} & \textbf{0.932} \\
\midrule
TeaCache~(\(\delta\)=0.2) & 64.4 & 4794 & 20.42 & 0.172 & 0.734 \\
TaylorSeer~(\(\mathcal{S}\)=3) & 68.8 & 5053 & 20.42 & 0.242 & 0.733 \\
SeaCache~(\(\delta\)=0.35) & \textbf{58.1} & \textbf{4598} & \textbf{26.46} & \textbf{0.133} & \textbf{0.857} \\
\bottomrule
\end{tabular}
\label{tab:main_hun}
\vspace{-5pt}
\end{table}

\begin{table}[t]
\centering
\setlength{\tabcolsep}{2pt}
\renewcommand{\arraystretch}{0.95}
\footnotesize
\caption{Quantitative comparison in \textit{Wan2.1 1.3B}~\cite{wan2025}.}
  \vspace{-5pt}
\begin{tabular}{lrrrrr}
\toprule
Method & Latency~(s) & TFLOPs & PSNR~\(\uparrow\) & LPIPS~\(\downarrow\) & SSIM~\(\uparrow\) \\
\midrule
Original (50 steps) & 176.3 & 8214 & -- & -- & -- \\
\midrule
TeaCache~(\(\delta\)=0.09) & 86.6 & 4107 & 20.84 & 0.171 & 0.721 \\
TaylorSeer~(\(\mathcal{S}\)=2) & 93.1 & 4189 & 16.15 & 0.336 & 0.543 \\
SeaCache~(\(\delta\)=0.2) & \textbf{83.9} & \textbf{3942} & \textbf{26.60} & \textbf{0.075} & \textbf{0.873} \\
\midrule
TeaCache~(\(\delta\)=0.15) & 63.6 & 2957 & 18.88 & 0.245 & 0.645 \\
TaylorSeer~(\(\mathcal{S}\)=3) & 67.1 & 2956 & 14.18 & 0.455 & 0.453 \\
SeaCache~(\(\delta\)=0.35) & \textbf{56.6} & \textbf{2793} & \textbf{21.78} & \textbf{0.170} & \textbf{0.740} \\
\bottomrule
\end{tabular}
\label{tab:main_wan}
\vspace{-5pt}
\end{table}

\begin{figure*}[t!]
  \centering
  \includegraphics[width=1.0\textwidth]{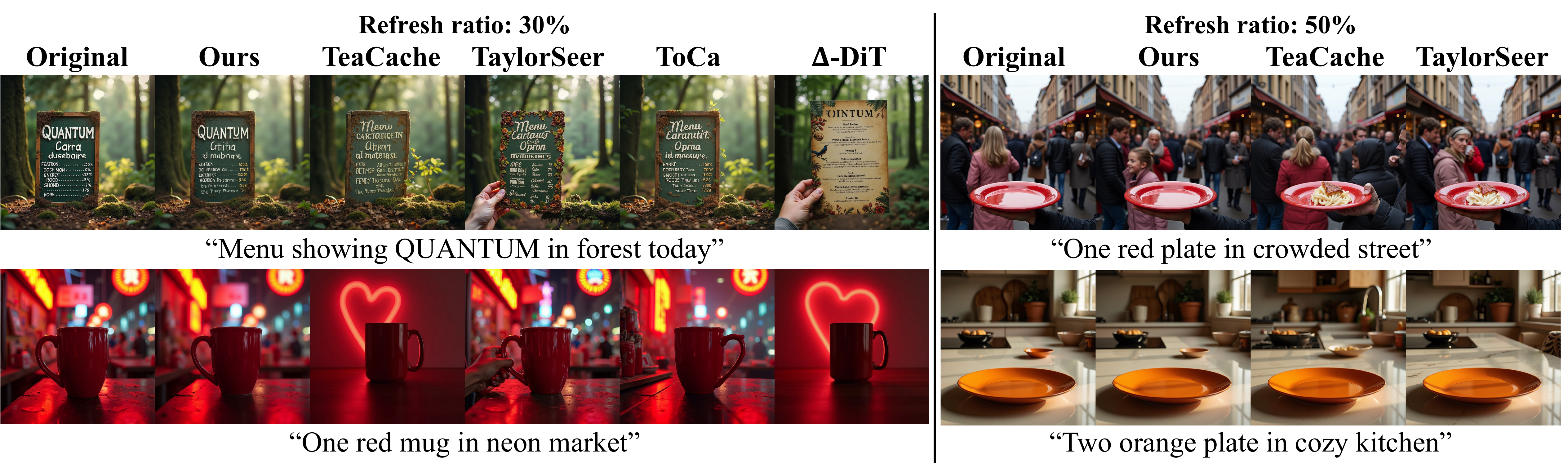}
  \vspace{-23pt}
  \caption{
  Qualitative comparison of SeaCache and baselines on \textit{FLUX} at refresh ratios of approximately \(30\%\) and \(50\%\).
  }
  \label{fig:qual_image}
  \vspace{-10pt}
\end{figure*}

\noindent\textbf{Text-to-video generation.}
For \textit{HunyuanVideo}~\cite{kong2024hunyuanvideo}, in Tab.~\ref{tab:main_hun}, SeaCache consistently achieves a stronger latency-quality trade-off than the baselines. In the higher cache budget setting (upper block), SeaCache reduces latency and TFLOPs while improving all metrics. PSNR increases by roughly 8\,dB over the strongest baseline, with lower LPIPS and higher SSIM. In the more aggressive setting (lower block), this trend remains. SeaCache runs faster than the baselines and still delivers clearly better overall metrics, whereas other methods show noticeable degradation.

For \textit{Wan2.1 1.3B}~\cite{wan2025}, a similar pattern appears in Tab.~\ref{tab:main_wan}. At the higher cache budget (upper block), SeaCache again reduces latency and TFLOPs while providing substantially higher PSNR and SSIM and lower LPIPS than TeaCache and TaylorSeer. Under the aggressive setting (lower block), SeaCache also shows the fastest latency and preserves reconstruction quality effectively, with consistently better metrics. Overall, these results indicate that the spectrum-aware schedule of SeaCache transfers well to video models and validate superior performance across architectures.

\subsection{Qualitative Comparison}
\noindent\textbf{Text-to-image generation.}
In Fig.~\ref{fig:qual_image}, we compare SeaCache with TeaCache~\cite{liu2025timestep} and TaylorSeer~\cite{liu2025reusing} at two cache budgets with refresh ratios of approximately \(30\%\) and \(50\%\).
SeaCache preserves both the semantic content and overall perceptual quality of the original images, whereas the baselines frequently lose text or fine details.
At a \(30\%\) refresh ratio, the baselines fail to reproduce the word ``quantum'' specified in the prompt, while SeaCache faithfully preserves the text present in the fully computed image.
At a \(50\%\) refresh ratio (second row), SeaCache generates two orange plates consistent with both the prompt and the original image, whereas the baselines either alter the plate color by filling the plates with food or remove one of the plates.

\begin{figure}[t]
  \centering
  \includegraphics[width=1\linewidth]{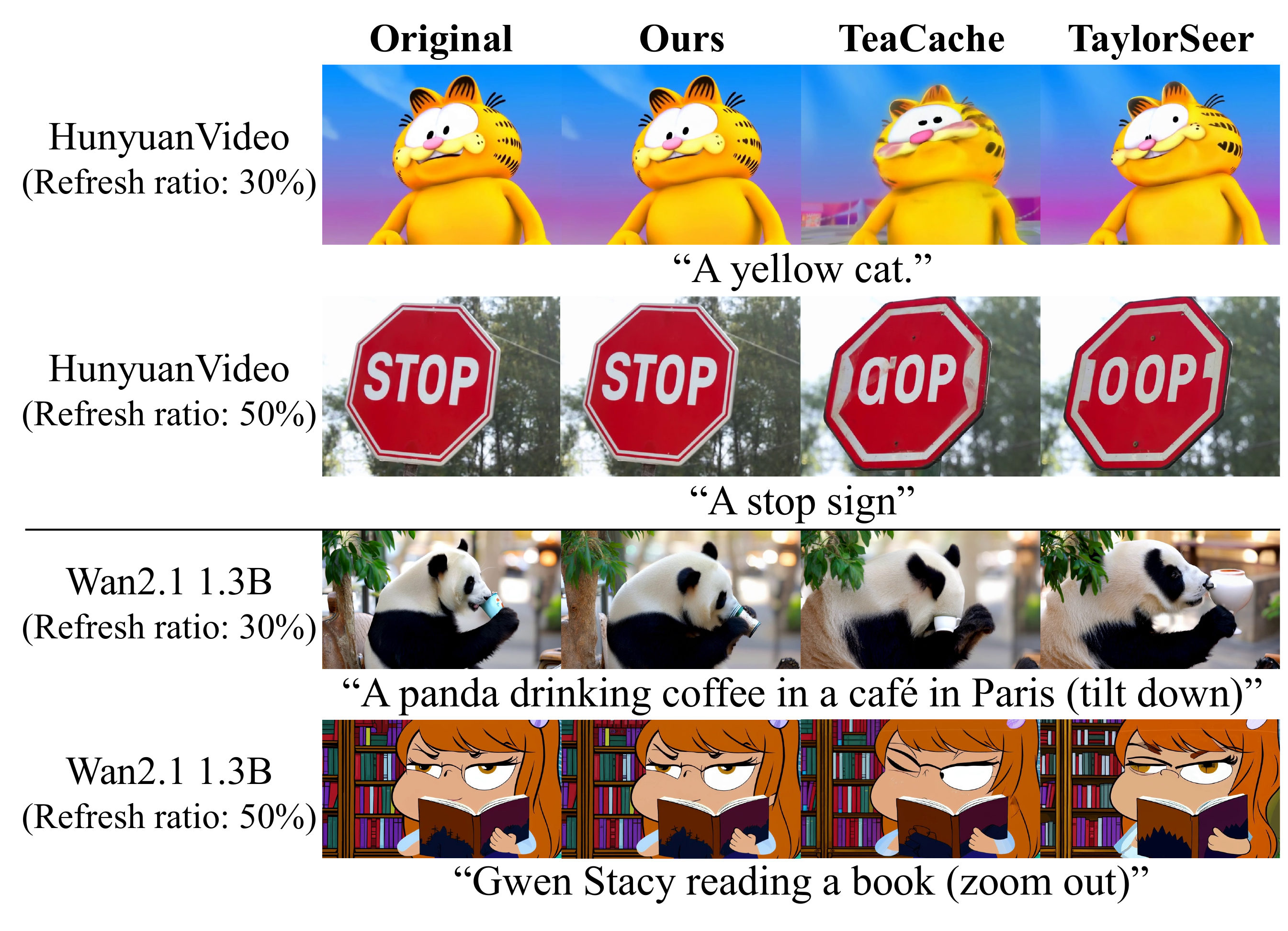}
  \vspace{-24pt}
  \caption{
    Qualitative comparison of text-to-video generative models at refresh ratios of approximately \(30\%\) and \(50\%\).
  }
  \label{fig:qual_video}
  \vspace{-10pt}
\end{figure}

\noindent\textbf{Text-to-video generation.}
We further conduct qualitative comparisons on text-to-video models \textit{HunyuanVideo} and \textit{Wan2.1 1.3B}, as shown in Fig.~\ref{fig:qual_video}.
For each prompt, we display the same frame index for the original model.
Our cache scheme preserves the content of the original implementation while using a comparable computation budget.
In the second row of \textit{HunyuanVideo}, SeaCache maintains a sharp and legible ``STOP'' sign that closely matches the original, whereas the baselines fail to synthesize the letters.
In the third row on \textit{Wan2.1 1.3B}, SeaCache produces clear pandas and coffee cups with fewer artifacts, while competing methods blur the foreground and introduce the distortions.

\subsection{Additional Analysis}

\begin{figure}[t]
  \centering
  \begin{subfigure}[t]{0.98\linewidth}
    \centering
    \includegraphics[width=0.98\linewidth]{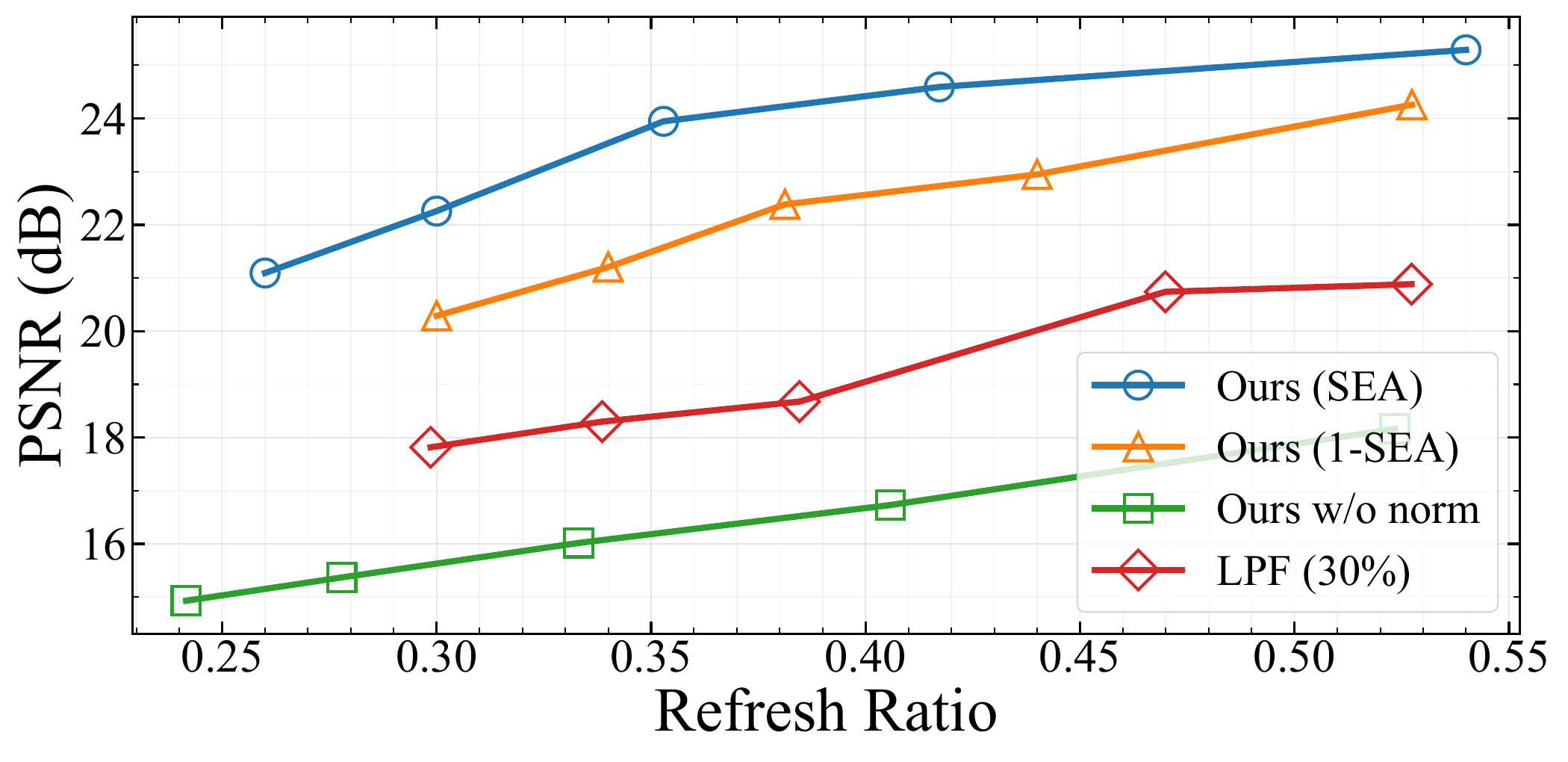}
    \vspace{-2pt}
    \caption{PSNR-refresh ratio trade-off on \textit{FLUX}~\cite{flux2024,labs2025flux1kontextflowmatching}.}
    \label{fig:ablation_flux}
  \end{subfigure}
  \begin{subfigure}[t]{0.98\linewidth}
    \centering
    \includegraphics[width=0.98\linewidth]{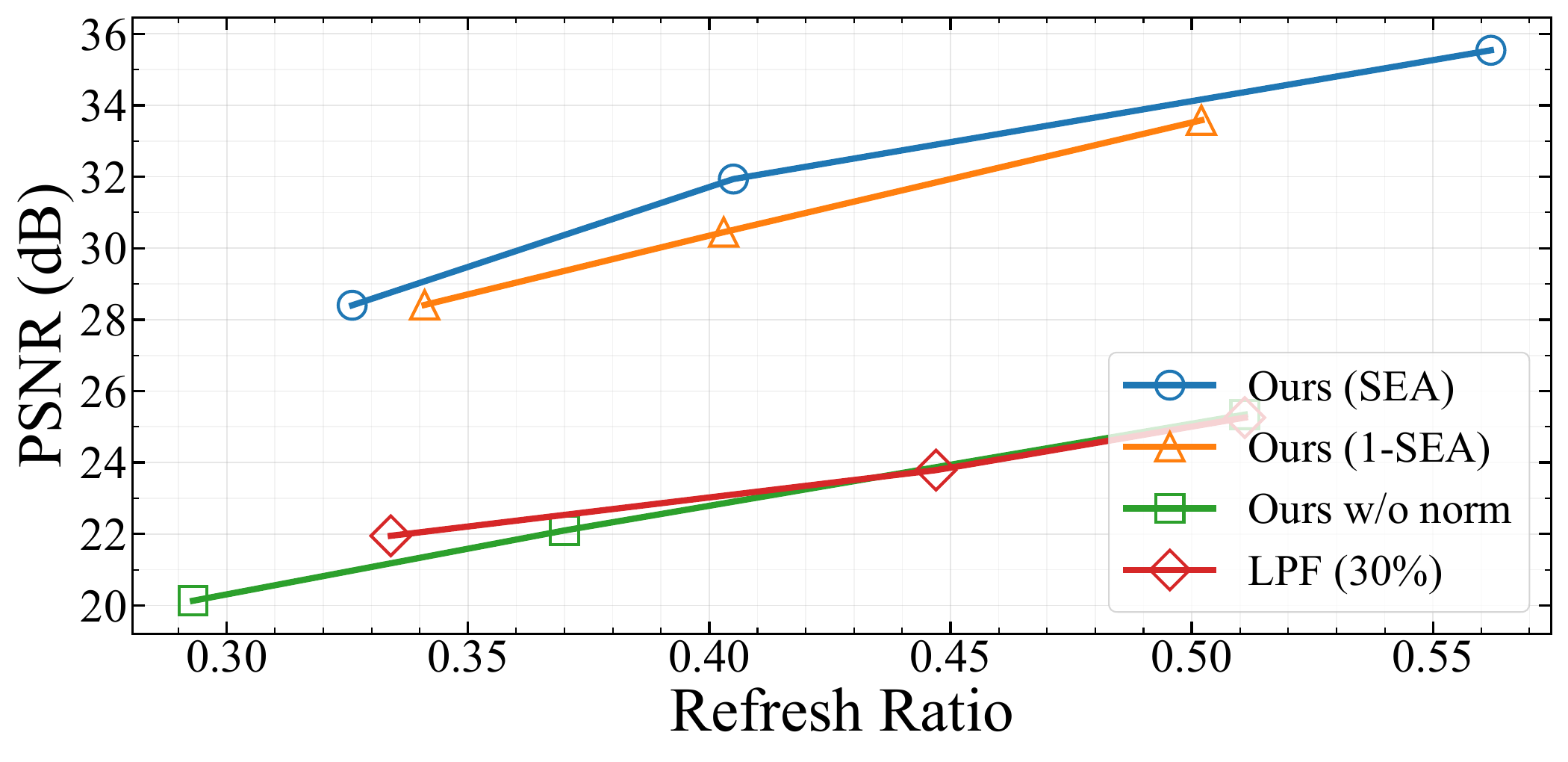}
    \vspace{-2pt}
    \caption{PSNR-refresh ratio trade-off on \textit{HunyuanVideo}~\cite{kong2024hunyuanvideo}.}
    \label{fig:ablation_hun}
  \end{subfigure}
  \vspace{-5pt}
  \caption{
  \textbf{Ablation on spectrum-aware filtering.}
  Trade-offs for different cache metrics on \textit{FLUX} and \textit{HunyuanVideo}. 
  Results are averaged over 200 prompts for \textit{FLUX} and 20 randomly selected from VBench for \textit{HunyuanVideo}, with the other settings fixed. 
  }
  \vspace{-4pt}
  \label{fig:ablation}
\end{figure}

\noindent\textbf{Ablation study.}
We quantitatively evaluate the effect of each design choice in SeaCache. 
Fig.~\ref{fig:ablation} compares four variants of the cache distance: the SEA filter, its complementary form \(1{-}\mathrm{SEA}\), a version without normalization, and a simple cutoff low-pass filter that only keeps the component of lowest \(30\%\) frequency. 
Across both \textit{FLUX} and \textit{HunyuanVideo}, the SEA filter shows the best PSNR-refresh ratio trade-off, while \(1{-}\mathrm{SEA}\) produces a similar but consistently lower curve, indicating that tracking the spectral evolution of the noise component is somewhat informative but less aligned with content redundancy than our signal-focused design. 
Removing normalization leads to a drop in PSNR, since the filtered feature magnitude drifts across timesteps and the cache metric becomes biased.
The static low-pass baseline (LPF 30\%) also performs noticeably worse than ours with SEA filter, showing that simply emphasizing low frequencies is insufficient and that the timestep-dependent spectral evolution captured by the SEA filter is crucial for effective cache scheduling.

\begin{figure}[t]
  \centering
  \includegraphics[width=0.98\linewidth]{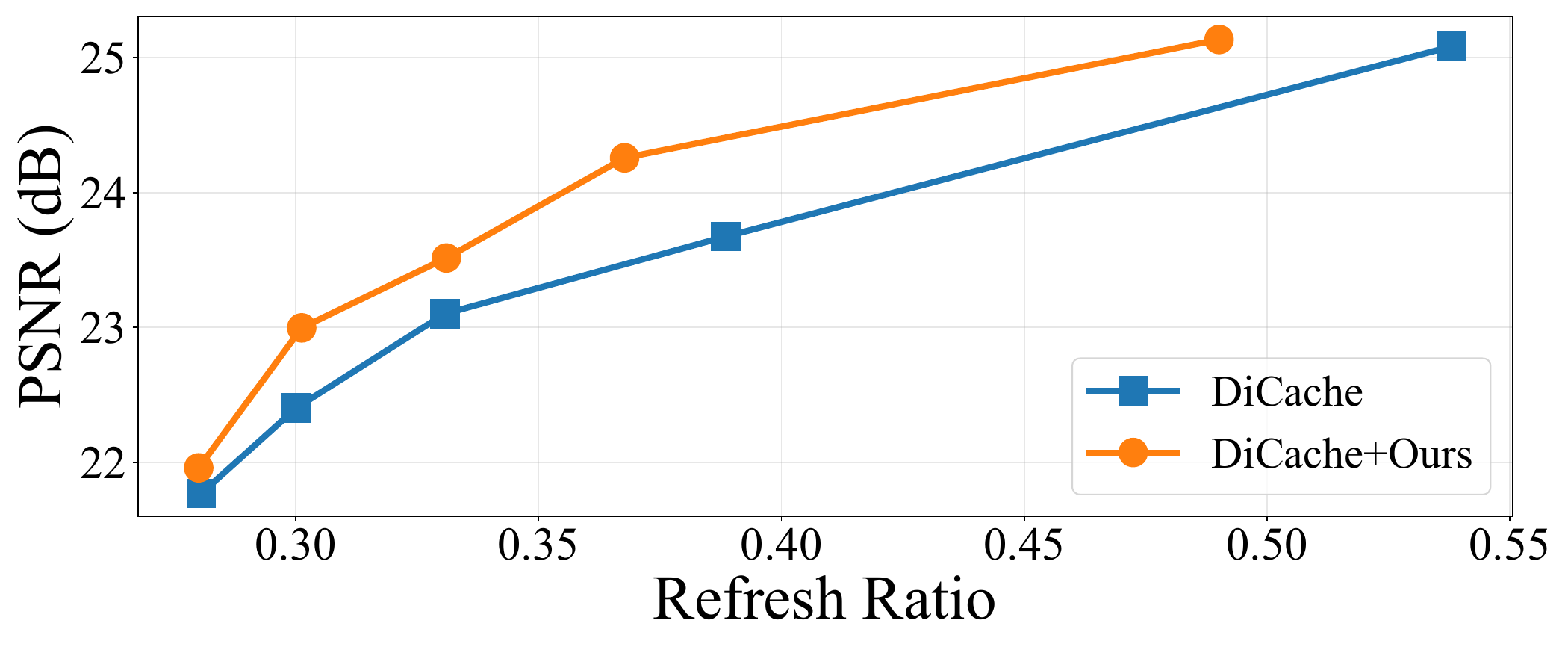}
  \vspace{-10pt}
  \caption{
  \textbf{Plug-and-play adaptation to DiCache.}
  PSNR-refresh ratio trade-off on \textit{FLUX} when applying the SEA-based cache metric to DiCache~\cite{bu2025dicache}.
  ``DiCache+Ours'' denotes DiCache combined with our SEA filter, while ``DiCache'' uses the original metric.}
  \label{fig:dicache_adapt}
  \vspace{-8pt}
\end{figure}

\noindent\textbf{Adaptation to other cache methods.}
To examine the plug-and-play nature of SeaCache, we integrate the SEA filter into DiCache~\cite{bu2025dicache}, a recent dynamic caching method that bases its metric on intermediate blocks. 
Instead of modifying the policy or network, we apply our SEA filtering to DiCache’s block-level features and reuse its original accumulation rule.
On \textit{FLUX}, the adapted variant (DiCache+Ours) achieves consistently higher PSNR for the same refresh ratio than the original DiCache, as shown in Fig.~\ref{fig:dicache_adapt}, while keeping latency and FLOPs comparable.
This result indicates that the proposed spectrum-aware distance is not limited to input-side features and can also enhance cache metrics defined on intermediate representations, suggesting broad compatibility with future caching schemes.

\begin{figure}[t]
  \centering
  \begin{subfigure}[t]{1\linewidth}
    \centering
    \vspace{-2.5pt}
    \includegraphics[width=1\linewidth, trim=0mm 120mm 70mm 0mm, clip]{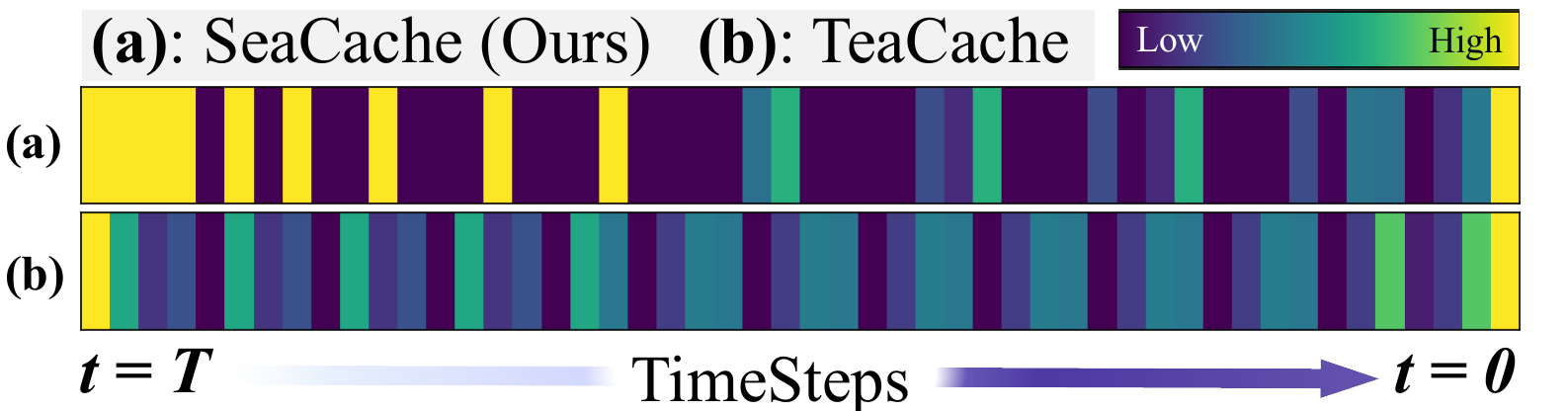}
    \label{fig:tmp}
  \end{subfigure}
  \vspace{-18pt}
  \caption{
  \textbf{Refresh pattern across timesteps on \textit{FLUX}.}
  Per-timestep refresh ratio at a \(30\%\) budget.
  (a)~SeaCache automatically concentrates refreshes on early timesteps, whereas (b)~TeaCache spreads refreshes more uniformly over the trajectory.
  }
  \label{fig:cache_ratio}
  \vspace{-8pt}
\end{figure}

\noindent\textbf{Cache ratio visualization.}
For each timestep in \textit{FLUX} with the 30\% refresh ratio setting, Fig.~\ref{fig:cache_ratio} shows the fraction of samples that trigger a refresh among 200 DrawBench prompts.
Bright cells indicate steps that are frequently computed, while dark cells correspond to steps that are almost always skipped. 
Many prior open-source methods~\cite{liu2025reusing,ma2024deepcache,bu2025dicache} fix several early steps to always compute in order to improve quality, introducing an extra hyperparameter that should be tuned by hand.
SeaCache instead concentrates most refreshes on early timesteps, aligned with the spectral-evolution prior, whereas TeaCache~\cite{liu2025timestep} distributes refreshes in a more gridlike pattern that does not adapt to timestep importance in Fig.~\ref{fig:cache_ratio}. 
This adaptive schedule removes the need to manually choose how many early steps to compute and uses the cache budget more effectively.


\section{Conclusion}
We study cache-based acceleration for diffusion models via spectral evolution, showing that raw feature distances used in prior cache-based approaches fail to separate the signal and the noise. We introduce SeaCache, a training-free policy that bases reuse decisions on a spectrally aligned space. From this analysis, we derived the Spectral-Evolution-Aware (SEA) filter, whose distances follow the full-compute trajectory more faithfully than unfiltered metrics.
By applying the SEA filter to input features, we obtain the schedules that adapt to content while respecting the spectral priors of the underlying diffusion model. We expect that incorporating spectral evolution into cache design can be combined with future acceleration methods.

\section*{Acknowledgments}
This work was partly supported by NAVER Cloud Corporation, MSIT/IITP (No. RS-2022-II220680, RS-2020-II201821, RS-2019-II190421, RS-2024-00459618, RS-2024-00360227, RS-2024-00437633, RS-2024-00437102, RS-2025-25442569), MSIT/NRF (No. RS-2024-00357729), and KNPA/KIPoT (No. RS-2025-25393280)
{
    \small
    \bibliographystyle{ieeenat_fullname}
    \bibliography{main}
}

\clearpage
\setcounter{page}{1}
\maketitlesupplementary


We first present the derivation analysis of the linear diffusion process.
We then provide additional experiments that further validate the effectiveness of the proposed SeaCache.

\section{Derivation of Optimal Linear Response}

To design a filter that reflects spectral evolution, we formalize how
the effective frequency band changes across timesteps.
Motivated by Spectral Diffusion~\cite{yang2023diffusion} and Wiener Filtering~\cite{wiener1964extrapolation}, we adopt the
timestep-dependent frequency response derived from the optimal linear
denoiser $h_t^\star$.
\footnote{Throughout this section, ``$\ast$'' denotes convolution in the frequency domain, the superscript ``${}^\star$'' denotes the optimal solution, and $\overline{(\cdot)}$ denotes complex conjugation of Fourier coefficients.}

\paragraph{Setup and assumptions.}
We consider the linear mixture of iterative denoising generative models
(DPMs and RFs) at timestep $t$,
\begin{equation}
x_t = a_t\,x_0 + b_t\,\epsilon, \qquad
\epsilon \sim \mathcal{N}(0,\mathbf{I}),
\label{eq:mix}
\end{equation}
where $x_0$ is the clean signal, assumed to be wide-sense stationary,
and $\epsilon$ is zero-mean white Gaussian noise with flat power
spectral density $S_\varepsilon(f)=1$.
We also assume that $x_0$ and $\epsilon$ are independent.

The Fourier-domain version of Eq.~\eqref{eq:mix} is
\begin{equation}
\mathcal{X}_t(f) = a_t\,\mathcal{X}_0(f) + b_t\,\mathcal{E}(f),
\label{eq:mix_fourier}
\end{equation}
where $\mathcal{X}_0(f)$, $\mathcal{X}_t(f)$, and $\mathcal{E}(f)$ are
the Fourier transforms of $x_0$, $x_t$ and $\epsilon$ at frequency $f$, respectively.

The filter $h_t$ estimates $x_0$ from $x_t$ as
\begin{equation}
\widehat{x}_0 = h_t \ast x_t
\quad\Longleftrightarrow\quad
\widehat{\mathcal{X}}_0(f) = \mathcal{H}_t(f)\,\mathcal{X}_t(f),
\label{eq:estimator}
\end{equation}
where $h_t$ is a linear reconstruction estimator,
$\mathcal{H}_t(f)$ is the frequency response of $h_t$,
and $\widehat{x}_0$, $\widehat{\mathcal{X}}_0(f)$ are the estimated
signal and its Fourier counterpart, respectively.

We define the signal reconstruction MSE objective, which is equivalent
to the denoising objective of diffusion models,
\begin{equation}
J_t = \big\|\,h_t \ast x_t - x_0\,\big\|_2^2,
\qquad
h_t^\star = \arg\min_{h_t}\;\mathbb{E}\!\left[\,J_t\,\right],
\label{eq:objective}
\end{equation}
where the expectation is taken over $(x_0,\epsilon)$.

\paragraph{Frequency-domain MSE expansion.}
By Parseval's theorem~\cite{mallat1999wavelet}, the reconstruction MSE (Eq.~\eqref{eq:objective}) decomposes as an integral over frequencies. Since $\mathcal{H}_t(f)$ acts independently at each frequency, minimizing the total MSE is equivalent to minimizing $J_t(f)$ for every $f$,
\begin{equation}
J_t(f) =
\mathbb{E}\!\left[\;\big|\;\mathcal{H}_t(f)\,\mathcal{X}_t(f)
- \mathcal{X}_0(f)\big|^2\;\right].
\label{eq:mse_freq}
\end{equation}

We now expand $J_t(f)$ using standard complex-valued quadratic
expansion:
{
\small
\begin{align}
J_t(f)
&= \mathbb{E}\!\left[\;\big|\;\mathcal{H}_{t}(f)\,\mathcal{X}_t(f)
- \mathcal{X}_0(f)\big|^2\;\right] \nonumber\\
&= \mathbb{E}\!\left[\;\big(\mathcal{H}_{t}(f)\,\mathcal{X}_t(f)
- \mathcal{X}_0(f)\big)\big(\overline{\mathcal{H}_{t}(f)}\,\overline{\mathcal{X}_t(f)}
- \overline{\mathcal{X}_0(f)}\big)\;\right] \nonumber\\
&= |\mathcal{H}_t(f)|^2\,\mathbb{E}\!\left[\,|\mathcal{X}_t(f)|^2\,\right]
- \mathcal{H}_t(f)\,\mathbb{E}\!\left[\,\mathcal{X}_t(f)\,\overline{\mathcal{X}_0(f)}\,\right] \nonumber\\
&\quad - \overline{\mathcal{H}_t(f)}\,\mathbb{E}\!\left[\,\mathcal{X}_0(f)\,\overline{\mathcal{X}_t(f)}\,\right]
+ \mathbb{E}\!\left[\,|\mathcal{X}_0(f)|^2\,\right],
\label{eq:expand}
\end{align}
}

where all quantities are evaluated at frequency $f$.

We next simplify the two expectation terms
$\mathbb{E}[\mathcal{X}_0(f)\,\overline{\mathcal{X}_t(f)}]$ and $\mathbb{E}[|\mathcal{X}_t(f)|^2]$, which will be used in the subsequent derivation.
Let $S_x(f)$ denote the power spectrum of $x_0$. The first term can be written as
{
\small
\begin{align}
\mathbb{E}\!\left[\mathcal{X}_0(f)\,\overline{\mathcal{X}_t(f)}\right]
&= \mathbb{E}\!\left[\mathcal{X}_0(f)\,\big(a_t\,\overline{\mathcal{X}_0(f)}
+ b_t\,\overline{\mathcal{E}(f)}\big)\right] \nonumber \\
&= a_t\,\mathbb{E}\!\left[|\mathcal{X}_0(f)|^2\right]
  + b_t\,\mathbb{E}\!\left[\mathcal{X}_0(f)\,\overline{\mathcal{E}(f)}\right] \nonumber \\
&= a_t\,\mathbb{E}\!\left[|\mathcal{X}_0(f)|^2\right] \nonumber \\
&= a_t\,S_x(f),
\label{eq:exp1}
\end{align}
}

since we assume that $x_0$ is wide-sense stationary and independent of the noise $\epsilon$, so $\mathbb{E}[\mathcal{X}_0(f)\,\overline{\mathcal{E}(f)}]=0$ in Eq.~\eqref{eq:exp1}.
Next, we expand the second expectation term:
{
\small
\begin{align}
\mathbb{E}\!\left[\,|\mathcal{X}_t(f)|^2\,\right]
&= \mathbb{E}\!\left[\big(a_t\mathcal{X}_0(f)+b_t\mathcal{E}(f)\big)
                    \big(a_t\,\overline{\mathcal{X}_0(f)}+b_t\,\overline{\mathcal{E}(f)}\big)\right] \nonumber \\
&= a_t^2\,\mathbb{E}\!\left[|\mathcal{X}_0(f)|^2\right]
 + b_t^2\,\mathbb{E}\!\left[|\mathcal{E}(f)|^2\right] \nonumber \\
&\quad\ + a_t b_t\,\mathbb{E}\!\left[\mathcal{X}_0(f)\,\overline{\mathcal{E}(f)}\right]
       + a_t b_t\,\mathbb{E}\!\left[\overline{\mathcal{X}_0(f)}\,\mathcal{E}(f)\right] \nonumber \\
&= a_t^2\,\mathbb{E}\!\left[|\mathcal{X}_0(f)|^2\right]
 + b_t^2\,\mathbb{E}\!\left[|\mathcal{E}(f)|^2\right] \nonumber \\
&= a_t^2\,S_x(f) + b_t^2\,S_\varepsilon(f) \nonumber \\
&= a_t^2\,S_x(f) + b_t^2,
\label{eq:exp2}
\end{align}
}
since in Eq.~\eqref{eq:exp2}, the cross terms vanish because
of independence,
$\mathbb{E}[\mathcal{X}_0(f)\overline{\mathcal{E}(f)}]
=\mathbb{E}[\overline{\mathcal{X}_0(f)} \mathcal{E}(f)]=0$, and whiteness of
the noise $\epsilon$ implies
$\mathbb{E}[|\mathcal{E}(f)|^2]=S_\varepsilon(f)=1$.

\paragraph{Optimality by differentiation.}
Differentiating Eq.~\eqref{eq:expand} with respect to
$\overline{\mathcal{H}_t(f)}$ using Wirtinger derivative~\cite{brandwood1983complex} and setting the result
to zero to find the optimal linear filter under the linear MMSE
criterion, we obtain
\begin{equation}
\frac{\partial J_t(f)}{\partial \overline{\mathcal{H}_t(f)}}
= \mathcal{H}_t(f)\,\mathbb{E}[\,|\mathcal{X}_t(f)|^2\,]
  - \mathbb{E}[\mathcal{X}_0(f)\,\overline{\mathcal{X}_t(f)}] = 0.
\label{eq:first_order}
\end{equation}
Using Eqs.~\eqref{eq:exp1} and \eqref{eq:exp2}, the unique minimizer is
\begin{equation}
\mathcal{H}_t^\star(f)
= \frac{\mathbb{E}[\mathcal{X}_0(f)\,\overline{\mathcal{X}_t(f)}]}
       {\mathbb{E}[\,|\mathcal{X}_t(f)|^2\,]}
= \frac{a_t\,S_x(f)}{a_t^2\,S_x(f) + b_t^2},
\label{eq:H_star}
\end{equation}
where $\mathcal{H}_t^\star(f)$ is the Fourier transform of $h^\star_t$.
We define the optimal frequency response
\begin{equation}
G_t(f) \;\triangleq\; \mathcal{H}_t^\star(f).
\label{eq:G_def}
\end{equation}

\paragraph{Power-law prior.}
We adopt an empirical natural-image power-law assumption
for the power spectrum~\cite{burton1987color,field1987relations,tolhurst1992amplitude,van1996modelling},
\begin{equation}
S_x(f) \simeq A\, |f|^{-\beta},
\end{equation}
where $A>0$ is an amplitude scaling factor and $\beta$ is a frequency
exponent. In our experiments, we set $A=1$ and $\beta=2$ for images and $\beta=3$ for videos.
Substituting this prior into the optimal response in
Eq.~\eqref{eq:H_star} gives
\begin{equation}
G_t(f)
= \frac{a_t\,|f|^{-\beta}}{a_t^2\,|f|^{-\beta}+b_t^2},
\end{equation}
which shows that the effective passband widens as $a_t$ increases
(\emph{spectral evolution}). Note that the SEA filter used in our method $G^{\text{norm}}_t(f)$ is a normalized variant of \(G_t(f)\). Its form is provided in the main manuscript.



\begin{table}[t]
  \caption{
  Runtime overhead of SEA filtering per sample, averaged over $10$ runs.
  }
\centering
\vspace{-5pt}
\setlength{\tabcolsep}{2pt}
\renewcommand{\arraystretch}{1.1}
\footnotesize
\begin{tabular}{|l|rrr|}
\hline
Model & SEA Filter~(s) & Latency~(s) & Overhead~(\%) \\
\hline
FLUX~(2D FFT) & 0.058 & 9.4 & 0.6 \\
HunyuanVideo~(3D FFT) & 0.362 & 90.8 & 0.4 \\
\hline
\end{tabular}
\label{tab:fft_overhead}
\end{table}

\begin{table}[t!]
\caption{
Runtime overhead of SEA filtering per sample on Wan2.1-14B-T2V at different output resolutions.
}
\centering
\vspace{-5pt}
\setlength{\tabcolsep}{2pt}
\renewcommand{\arraystretch}{0.95}
\footnotesize
\begin{tabular}{|l|rrr|}
\hline
Resolution & SEA Filter~(s) & Total Latency~(s) & Overhead~(\%) \\
\hline
480p & 0.668 & 161.5 & 0.41 \\
720p & 1.539 & 561.1 & 0.27 \\
\hline
\end{tabular}
\label{tab:sea_overhead}
\vspace{-5pt}
\end{table}

\begin{table*}[t]
\vspace{15pt}
\centering
\setlength{\tabcolsep}{4pt}
\small
\caption{VBench metrics in \textit{HunyuanVideo}.}
\begin{tabular}{lcccccccc}
\toprule
Models &
\begin{tabular}[c]{@{}c@{}}Subject\\Consistency\end{tabular} &
\begin{tabular}[c]{@{}c@{}}Background\\Consistency\end{tabular} &
\begin{tabular}[c]{@{}c@{}}Temporal\\Flickering\end{tabular} &
\begin{tabular}[c]{@{}c@{}}Motion\\Smoothness\end{tabular} &
\begin{tabular}[c]{@{}c@{}}Dynamic\\Degree\end{tabular} &
\begin{tabular}[c]{@{}c@{}}Aesthetic\\Quality\end{tabular} &
\begin{tabular}[c]{@{}c@{}}Imaging\\Quality\end{tabular} &
\begin{tabular}[c]{@{}c@{}}Object\\Class\end{tabular} \\
\midrule
TeaCache ($\delta$=0.12)      & 95.59\% & 95.99\% & \underline{99.14\%} & 98.77\% & \underline{62.50\%} & \underline{60.92\%} & \underline{62.07\%} & \textbf{86.31\%} \\
TaylorSeer (\(\mathcal{S}\)=2) & \underline{95.75\%} & \underline{96.20\%} & 99.09\% & \underline{98.83\%} & \textbf{63.89\%} & \textbf{60.93\%} & \textbf{62.73\%} & 83.47\% \\
SeaCache ($\delta$=0.19)      & \textbf{95.77\%} & \textbf{96.28\%} & \textbf{99.15\%} & \textbf{98.88\%} & 62.50\% & 60.55\% & 62.01\% & \underline{85.28\%} \\
\midrule
TeaCache ($\delta$=0.2)      & 95.57\% & 96.04\% & \underline{99.18\%} & 98.76\% & \underline{62.50\%} & \underline{60.28\%} & 60.28\% & \textbf{86.47\%} \\
TaylorSeer (\(\mathcal{S}\)=3) & \underline{95.67\%} & \underline{96.18\%} & 99.07\% & \underline{98.86\%} & \textbf{63.89\%} & \textbf{60.64\%} & \textbf{63.25\%} & 82.20\% \\
SeaCache ($\delta$=0.35)     & \textbf{95.78\%} & \textbf{96.35\%} & \textbf{99.20\%} & \textbf{98.92\%} & 61.11\% & 60.00\% & \underline{61.02\%} & \underline{82.59\%} \\
\midrule\midrule
Models &
\begin{tabular}[c]{@{}c@{}}Multiple\\Objects\end{tabular} &
\begin{tabular}[c]{@{}c@{}}Human\\Action\end{tabular} &
Color &
\begin{tabular}[c]{@{}c@{}}Spatial\\Relationship\end{tabular} &
Scene &
\begin{tabular}[c]{@{}c@{}}Temporal\\Style\end{tabular} &
\begin{tabular}[c]{@{}c@{}}Appearance\\Style\end{tabular} &
\begin{tabular}[c]{@{}c@{}}Overall\\Consistency\end{tabular} \\
\midrule
TeaCache ($\delta$=0.12)      & \textbf{64.71\%} & \textbf{96.00\%} & 89.61\% & \underline{61.84\%} & \textbf{42.81\%} & 24.39\% & \underline{19.85\%} & \textbf{26.91\%} \\
TaylorSeer (\(\mathcal{S}\)=2) & 58.38\% & \underline{95.00\%} & \textbf{90.87\%} & 60.80\% & 40.48\% & \underline{24.44\%} & \textbf{19.89\%} & 26.60\% \\
SeaCache ($\delta$=0.19)      & \underline{63.64\%} & 94.00\% & \underline{90.26\%} & \textbf{62.96\%} & \underline{40.92\%} & \textbf{24.66\%} & 19.83\% & \underline{26.63\%} \\
\midrule
TeaCache ($\delta$=0.2)      & \textbf{63.34\%} & \underline{92.00\%} & \underline{89.81\%} & \underline{59.65\%} & \textbf{44.48\%} & 24.26\% & 19.93\% & \textbf{26.68\%} \\
TaylorSeer (\(\mathcal{S}\)=3) & \underline{60.06\%} & \underline{92.00\%} & 89.26\% & 57.78\% & 41.72\% & \textbf{24.35\%} & \underline{20.02\%} & \underline{26.57\%} \\
SeaCache ($\delta$=0.35)     & 58.38\% & \textbf{94.00\%} & \textbf{92.24\%} & \textbf{60.63\%} & \underline{42.88\%} & \underline{24.34\%} & \textbf{20.10\%} & 26.33\% \\
\bottomrule
\end{tabular}
\label{tab:vbench-hun}
\end{table*}

\begin{table*}[t]
\centering
\vspace{10pt}
\setlength{\tabcolsep}{4pt}
\small
\caption{VBench metrics in \textit{Wan2.1~1.3B}.}
\begin{tabular}{lcccccccc}
\toprule
Models &
\begin{tabular}[c]{@{}c@{}}Subject\\Consistency\end{tabular} &
\begin{tabular}[c]{@{}c@{}}Background\\Consistency\end{tabular} &
\begin{tabular}[c]{@{}c@{}}Temporal\\Flickering\end{tabular} &
\begin{tabular}[c]{@{}c@{}}Motion\\Smoothness\end{tabular} &
\begin{tabular}[c]{@{}c@{}}Dynamic\\Degree\end{tabular} &
\begin{tabular}[c]{@{}c@{}}Aesthetic\\Quality\end{tabular} &
\begin{tabular}[c]{@{}c@{}}Imaging\\Quality\end{tabular} &
\begin{tabular}[c]{@{}c@{}}Object\\Class\end{tabular} \\
\midrule
TeaCache ($\delta$=0.09)      & \underline{95.89\%} & \textbf{97.09\%} & \underline{98.30\%} & 97.37\% & 81.94\% & \textbf{62.48\%} & 67.88\% & 80.46\% \\
TaylorSeer (\(\mathcal{S}\)=2) & 95.78\% & 96.90\% & \textbf{98.37\%} & \textbf{97.47\%} & \textbf{88.89\%} & 62.14\% & \textbf{68.08\%} & \textbf{82.75\%} \\
SeaCache ($\delta$=0.2)       & \textbf{95.96\%} & \underline{97.05\%} & 98.20\% & \underline{97.41\%} & \underline{84.72\%} & \underline{62.31\%} & \underline{68.01\%} & \underline{81.17\%} \\
\midrule
TeaCache ($\delta$=0.15)      & \textbf{96.04\%} & \textbf{97.02\%} & \textbf{98.21\%} & 97.35\% & \underline{83.33\%} & \textbf{62.25\%} & 67.47\% & \textbf{80.22\%} \\
TaylorSeer (\(\mathcal{S}\)=3) & 95.32\% & 96.54\% & \textbf{98.21\%} & \textbf{97.48\%} & \textbf{84.72\%} & 60.85\% & \textbf{67.83\%} & 78.32\% \\
SeaCache ($\delta$=0.35)      & \underline{96.03\%} & \underline{97.00\%} & 98.12\% & \underline{97.39\%} & 81.94\% & \underline{61.71\%} & \underline{67.66\%} & \underline{79.75\%} \\
\midrule\midrule
Models &
\begin{tabular}[c]{@{}c@{}}Multiple\\Objects\end{tabular} &
\begin{tabular}[c]{@{}c@{}}Human\\Action\end{tabular} &
Color &
\begin{tabular}[c]{@{}c@{}}Spatial\\Relationship\end{tabular} &
Scene &
\begin{tabular}[c]{@{}c@{}}Temporal\\Style\end{tabular} &
\begin{tabular}[c]{@{}c@{}}Appearance\\Style\end{tabular} &
\begin{tabular}[c]{@{}c@{}}Overall\\Consistency\end{tabular} \\
\midrule
TeaCache ($\delta$=0.09)      & 52.67\% & \textbf{72.00\%} & \underline{92.95\%} & \underline{71.46\%} & \underline{23.91\%} & \underline{23.07\%} & \underline{20.06\%} & \underline{23.42\%} \\
TaylorSeer (\(\mathcal{S}\)=2) & \underline{53.73\%} & \underline{70.00\%} & 91.22\% & \textbf{75.48\%} & \textbf{30.09\%} & 22.75\% & \textbf{20.13\%} & 23.41\% \\
SeaCache ($\delta$=0.2)       & \textbf{53.89\%} & \underline{70.00\%} & \textbf{93.01\%} & 69.50\% & 22.89\% & \textbf{23.32\%} & 20.04\% & \textbf{23.51\%} \\
\midrule
TeaCache ($\delta$=0.15)      & \underline{51.91\%} & \textbf{72.00\%} & \textbf{90.56\%} & \underline{67.67\%} & \textbf{24.27\%} & \textbf{22.98\%} & \underline{20.09\%} & \textbf{23.58\%} \\
TaylorSeer (\(\mathcal{S}\)=3) & 45.05\% & 69.00\% & 87.83\% & 60.79\% & 20.20\% & 22.37\% & \textbf{20.64\%} & 23.17\% \\
SeaCache ($\delta$=0.35)      & \textbf{53.20\%} & 68.00\% & \underline{89.67\%} & \textbf{69.57\%} & \underline{23.62\%} & \underline{22.96\%} & 20.06\% & \underline{23.18\%} \\
\bottomrule
\end{tabular}
\label{tab:vbench-wan}
\end{table*}

\begin{table}[t]
\vspace{10pt}
\caption{Comparison of avg. rank on VBench in \textit{HunyuanVideo}.}
\centering
\setlength{\tabcolsep}{4pt}
\small
\begin{tabular}{l c | l c}
\toprule
Method (\(\approx50\%\)) & Rank~\(\downarrow\) &
Method (\(\approx30\%\)) & Rank~\(\downarrow\) \\
\midrule
TeaCache (\(\delta\)=0.12)      & \underline{2.03} & TeaCache (\(\delta\)=0.20)      & 2.16 \\
TaylorSeer (\(\mathcal{S}\)=2)  & 2.06             & TaylorSeer (\(\mathcal{S}\)=3)  & \underline{2.09} \\
SeaCache (\(\delta\)=0.19)      & \textbf{1.91}    & SeaCache (\(\delta\)=0.35)      & \textbf{1.75} \\
\bottomrule
\end{tabular}
\label{tab:avg-rank-hunyuan}
\end{table}

\begin{table}[t]
\vspace{10pt}
\caption{Comparison of avg. rank on VBench in \textit{Wan2.1~1.3B}.}
\centering
\setlength{\tabcolsep}{4pt}
\small
\begin{tabular}{l c | l c}
\toprule
Method (\(\approx50\%\)) & Rank~\(\downarrow\) &
Method (\(\approx30\%\)) & Rank~\(\downarrow\) \\
\midrule
TeaCache (\(\delta\)=0.09)      & 2.13             & TeaCache (\(\delta\)=0.15)      & \textbf{1.53} \\
TaylorSeer (\(\mathcal{S}\)=2)  & \textbf{1.91}    & TaylorSeer (\(\mathcal{S}\)=3)  & 2.34 \\
SeaCache (\(\delta\)=0.30)      & \underline{1.97} & SeaCache (\(\delta\)=0.35)      & \underline{2.13} \\
\bottomrule
\end{tabular}
\label{tab:avg-rank-wan}
\end{table}

\begin{table*}[t]
\vspace{20pt}
\caption{CompressedVQA~\cite{sun2021deep} scores on \textit{HunyuanVideo} and \textit{Wan2.1~1.3B} under single-scale and multi-scale settings.}
\centering
\small
\begin{tabular}{lcc|ccc}
\hline
\multicolumn{3}{c|}{\textit{HunyuanVideo}} &
\multicolumn{3}{c}{\textit{Wan2.1~1.3B}} \\
\hline
Method (\(\approx50\%\)) & Single-scale score~\(\uparrow\) & Multi-scale score~\(\uparrow\) &
Method (\(\approx50\%\)) & Single-scale score~\(\uparrow\) & Multi-scale score~\(\uparrow\) \\
\hline
TeaCache (\(\delta\)=0.12)      & 2.72 & 2.76 & TeaCache (\(\delta\)=0.09)      & 2.97 & 3.03 \\
TaylorSeer (\(\mathcal{S}\)=2)  & 2.92 & 2.95 & TaylorSeer (\(\mathcal{S}\)=2)  & 1.90 & 1.95 \\
SeaCache (\(\delta\)=0.19)      & \textbf{3.98} & \textbf{3.99} & SeaCache (\(\delta\)=0.30)      & \textbf{3.93} & \textbf{3.95} \\
\hline
Method (\(\approx30\%\)) & Single-scale score~\(\uparrow\) & Multi-scale score~\(\uparrow\) &
Method (\(\approx30\%\)) & Single-scale score~\(\uparrow\) & Multi-scale score~\(\uparrow\) \\
\hline
TeaCache (\(\delta\)=0.20)      & 2.11 & 2.16 & TeaCache (\(\delta\)=0.15)      & 2.44 & 2.49 \\
TaylorSeer (\(\mathcal{S}\)=3)  & 2.22 & 2.26 & TaylorSeer (\(\mathcal{S}\)=3)  & 1.38 & 1.42 \\
SeaCache (\(\delta\)=0.35)      & \textbf{3.13} & \textbf{3.17} & SeaCache (\(\delta\)=0.35)      & \textbf{3.09} & \textbf{3.11} \\
\hline
\end{tabular}
\label{tab:compressedvqa}
\vspace{1pt}
\end{table*}

\section{Runtime Overhead of SEA Filtering}
\label{sec:fft_overhead}
At every sampling step, SeaCache inserts an additional FFT $\rightarrow$ frequency-domain filtering $\rightarrow$ iFFT pass to construct SEA-filtered features. Thus, we measure how much of the end-to-end sampling time this pass occupies under a 50\% caching ratio, keeping all other settings identical to the main experiments.
For \textit{FLUX}~\cite{flux2024,labs2025flux1kontextflowmatching} with SeaCache, the SEA filtering pass takes on average $0.058$\,s per sample out of a total latency of $9.4$\,s, corresponding to only about $0.6$\% of the overall generation time. For HunyuanVideo~\cite{kong2024hunyuanvideo} with SeaCache, the 3D FFT-based SEA filtering costs $0.362$\,s per sample while the total latency is $90.8$\,s, roughly $0.4$\% of the end-to-end runtime. As summarized in Tab.~\ref{tab:fft_overhead}, the SEA filtering introduces a negligible runtime overhead while enabling substantially better preservation of the original outputs compared to prior caching schemes.

To further quantify the SEA filter and FFT/iFFT overhead on a large text-to-video~(T2V) diffusion model, we additionally profile LightX2V~\cite{lightx2v} on Wan2.1-14B-T2V~\cite{wan2025} under a 50\% caching ratio, using a single Blackwell Pro~6000 GPU, while keeping all other settings identical.
Since sampling is performed in a compressed latent space, the SEA filtering pass occupies only a tiny fraction of the end-to-end runtime.
As shown in Tab.~\ref{tab:sea_overhead}, the overhead stays below 1\% in practice and remains small even when increasing the output resolution (0.41\% at 480p and 0.27\% at 720p).

\section{Compatibility with Fast Inference Works}
\label{sec:compat_fast}

We additionally validate SeaCache on Wan2.1-T2V under two orthogonal acceleration settings: (i) a distilled sampler (LightX2V~\cite{lightx2v}, 16-step) and (ii) an efficient-attention variant (Jenga~\cite{zhang2025training}, 50-step), using each method's default configuration.
All results are evaluated on VBench~\cite{huang2024vbench} at 480p with 41 frames, using videos generated from 50 randomly sampled VBench prompts.
Under comparable refresh ratio budgets, SeaCache consistently improves quality over TeaCache and vanilla step reduction, as shown in Tab.~\ref{tab:distill_eff}.

\begin{table}[t]
\centering
\setlength{\tabcolsep}{3.5pt}
\renewcommand{\arraystretch}{0.90}
\footnotesize
\caption{Compatibility with fast inference works on Wan2.1-T2V: LightX2V~\cite{lightx2v} (14B, 16-step) and Jenga~\cite{zhang2025training} (1.3B, 50-step), evaluated under comparable refresh ratio budgets.}
\vspace{-5pt}
\begin{tabular}{clrrrr}
\toprule
 & Method & Refresh Ratio & PSNR & LPIPS & SSIM \\
\midrule
\multirow{3}{*}{\makecell[c]{\textbf{Distillation}\\(LightX2V~\cite{lightx2v})}} 
& Vanilla  & 25\% (4 steps)  & 11.444 & 0.475 & 0.405 \\
& TeaCache & 25\% (4 steps)  & 11.762 & 0.480 & 0.420 \\
& SeaCache & 25\% (4 steps)  & \textbf{11.926} & \textbf{0.465} & \textbf{0.432} \\
\midrule
\multirow{6}{*}{\makecell[c]{\textbf{Efficient Attn.}\\(Jenga~\cite{zhang2025training})}} 
& Vanilla  & 50\% (25 steps) & 15.154 & 0.357 & 0.604 \\
& TeaCache & 50\% (25 steps) & 19.463 & 0.191 & 0.744 \\
& SeaCache & 48\% (24 steps) & \textbf{24.453} & \textbf{0.097} & \textbf{0.852} \\
\cmidrule(lr){2-6}
& Vanilla  & 32\% (16 steps) & 13.440 & 0.455 & 0.534 \\
& TeaCache & 32\% (16 steps) & 17.692 & 0.259 & 0.681 \\
& SeaCache & 32\% (16 steps) & \textbf{20.259} & \textbf{0.194} & \textbf{0.748} \\
\bottomrule
\end{tabular}
\label{tab:distill_eff}
\end{table}

\section{Additional Evaluation}

\subsection{Quantitative Comparison in T2V Generation}
\label{sec:vbench}

\noindent\textbf{VBench on \textit{HunyuanVideo}.}
We evaluate SeaCache against TeaCache~\cite{liu2025timestep} and TaylorSeer~\cite{guan2025forecasting} on all VBench~\cite{huang2024vbench} dimensions (Tab.~\ref{tab:vbench-hun}), where the upper rows correspond to the $50$\% refresh-ratio budget and the lower rows to the $30$\% budget.
All detailed settings follow the main manuscript.
Aggregating by average rank across dimensions (Tab.~\ref{tab:avg-rank-hunyuan}), SeaCache ranks first under both budgets, scoring \(1.91\) vs.\ \(2.03/2.06\) at \(\approx50\%\), and \(1.75\) vs.\ \(2.16/2.09\) at \(\approx30\%\).
This indicates the strongest overall performance across VBench dimensions on \textit{HunyuanVideo}.

\noindent\textbf{VBench on \textit{Wan2.1~1.3B}.}
We repeat the evaluation on all VBench dimensions for \textit{Wan2.1}~\cite{wan2025} (Tab.~\ref{tab:vbench-wan}) with the same two budgets and the same experimental details as in the main manuscript.
In aggregate (Tab.~\ref{tab:avg-rank-wan}), SeaCache delivers stable performance across dimensions, ranking second under both budgets, \(1.97\) at \(\approx50\%\) (vs.\ the best \(1.91\)) and \(2.13\) at \(\approx30\%\) (vs.\ the best \(1.53\)).
Although our cache configurations are designed to closely track the original full-refresh sampling trajectory, the VBench results on \textit{Wan2.1} still show that SeaCache provides robust performance across dimensions and refresh-ratio budgets.

\noindent\textbf{CompressedVQA on T2V.}
To further quantify how caching affects video quality, we report scores from CompressedVQA~\cite{sun2021deep}, a full-reference video quality assessment (VQA) metric.
For each video, we treat the uncached trajectory as the reference and compute single-scale and multi-scale scores between the cached outputs.
Tab.~\ref{tab:compressedvqa} summarizes the results on \textit{HunyuanVideo} and \textit{Wan2.1~1.3B} at two cache budgets with refresh ratios of approximately \(50\%\) and \(30\%\).
Across both models and budgets, SeaCache consistently achieves the highest single-scale and multi-scale scores among all caching baselines, indicating that it best preserves the visual quality of the original trajectory while still enjoying substantial reductions in the refresh ratio.

\subsection{Comparison with MagCache}
\label{sec:magcache}

We further compare SeaCache with MagCache~\cite{ma2025magcache} under matched refresh ratios by tuning the cache threshold \(\delta\).
We use the default MagCache configuration and report results.
For \textit{FLUX.1-dev}, we follow the manuscript protocol (DrawBench, 200 prompts).
For \textit{Wan2.1~1.3B} T2V, we evaluate at 480p with 41 frames using 50 randomly sampled prompts from VBench~\cite{huang2024vbench}.
As shown in Tab.~\ref{tab:mag_vs_sea_rr_combined}, SeaCache consistently improves quality at the same refresh ratio.
We attribute this to SeaCache's input-adaptive redundancy estimation, whereas MagCache relies on a fixed magnitude threshold, which is less responsive to content- and timestep-dependent variations.

\begin{table*}
\centering
\small
\setlength{\tabcolsep}{4pt}
\caption{Comparison with MagCache at matched refresh ratios (R.R.) by tuning the cache threshold \(\delta\).
We report full-reference quality against the uncached outputs on \emph{FLUX.1-dev} and \emph{Wan2.1~1.3B}-T2V.
SeaCache shows higher PSNR and lower LPIPS than MagCache~\cite{ma2025magcache} at the same refresh ratio.}
\vspace{-5pt}
\begin{tabular}{lrrr|lrrr}
\midrule
Method & R.R. & PSNR~$\uparrow$ & LPIPS~$\downarrow$ & Method & R.R. & PSNR~$\uparrow$ & LPIPS~$\downarrow$ \\
\midrule
\multicolumn{8}{c}{\emph{FLUX.1-dev} (50-step)} \\
\midrule
MagCache~(\(\delta\)=0.04)~\cite{ma2025magcache} & 52\% & 29.96 & 0.056 &
MagCache~(\(\delta\)=0.15)~\cite{ma2025magcache} & 34\% & 24.73 & 0.126 \\
SeaCache~(\(\delta\)=0.215) & 52\% & \textbf{30.37} & \textbf{0.053} &
SeaCache~(\(\delta\)=0.4) & 34\% & \textbf{24.97} & \textbf{0.123} \\
\midrule
MagCache~(\(\delta\)=0.07)~\cite{ma2025magcache} & 44\% & 27.89 & 0.079 &
MagCache~(\(\delta\)=0.35)~\cite{ma2025magcache} & 28\% & 22.51 & 0.179 \\
SeaCache~(\(\delta\)=0.27) & 44\% & \textbf{28.09} & \textbf{0.072} &
SeaCache~(\(\delta\)=0.55) & 28\% & \textbf{23.01} & \textbf{0.172} \\
\midrule
\multicolumn{8}{c}{\emph{Wan2.1 1.3B T2V} (50-step)} \\
\midrule
MagCache~(\(\delta\)=0.055)~\cite{ma2025magcache} & 50\% & 25.55 & 0.079 &
MagCache~(\(\delta\)=0.15)~\cite{ma2025magcache} & 32\% & 19.32 & 0.226 \\
SeaCache~(\(\delta\)=0.19) & 49\% & \textbf{29.55} & \textbf{0.047} &
SeaCache~(\(\delta\)=0.4) & 32\% & \textbf{21.98} & \textbf{0.156} \\
\hline
\end{tabular}
\label{tab:mag_vs_sea_rr_combined}
\end{table*}

\begin{figure*}[t]
\includegraphics[width=0.9\textwidth]{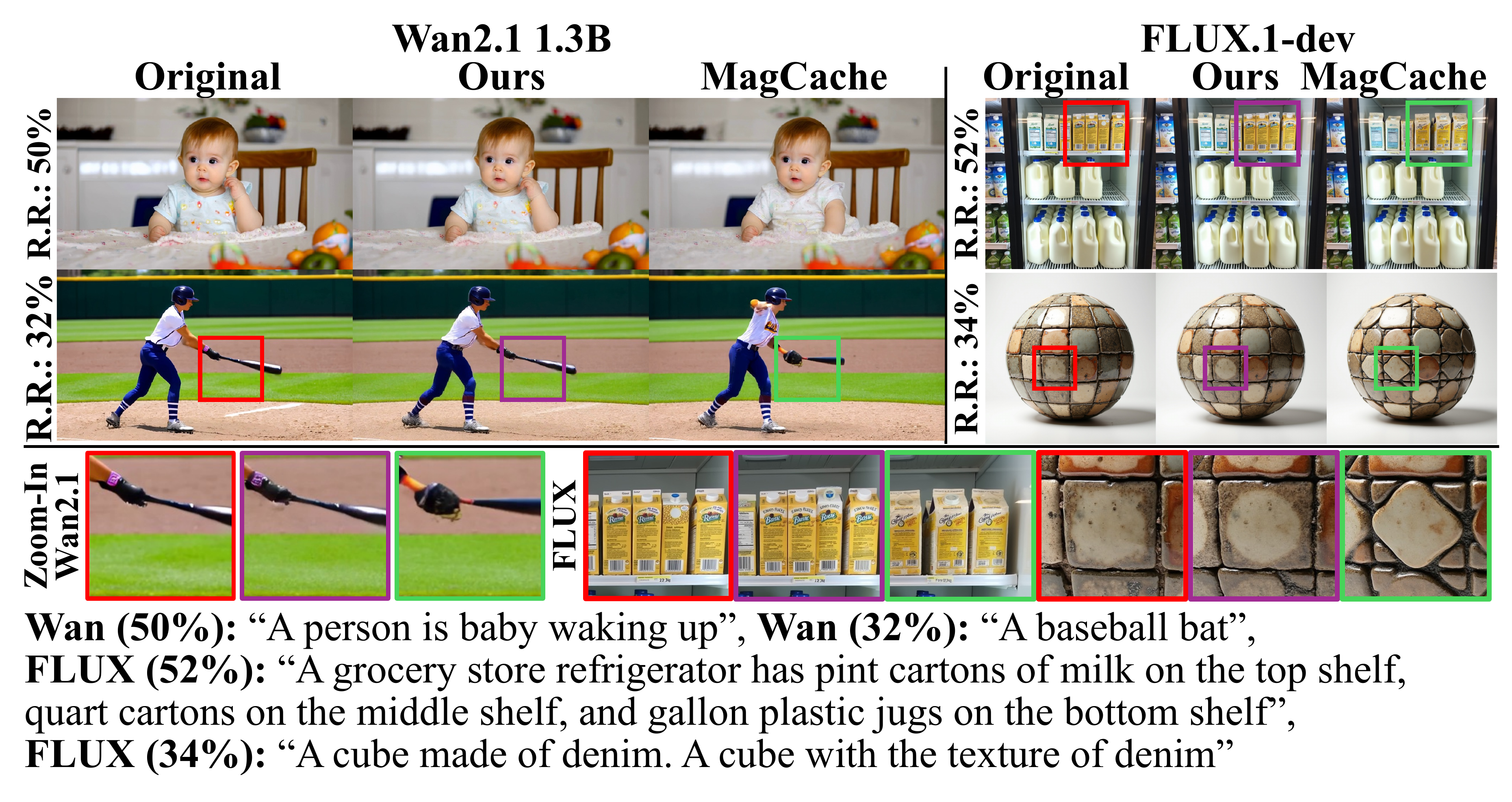}\centering
\vspace{-10pt}
\caption{Qualitative comparison between MagCache and SeaCache at matched refresh ratios on \emph{FLUX.1-dev} and \emph{Wan2.1~1.3B} T2V.
At the same refresh ratio, SeaCache better preserves the uncached trajectory while reducing refresh operations.}
\vspace{-0.4cm}
\label{fig:reb_qual}
\end{figure*}

\subsection{Qualitative Comparison in T2I Generation}
\begin{figure*}[t!]
  \centering
  \vspace{10pt}
  \includegraphics[width=1.0\textwidth]{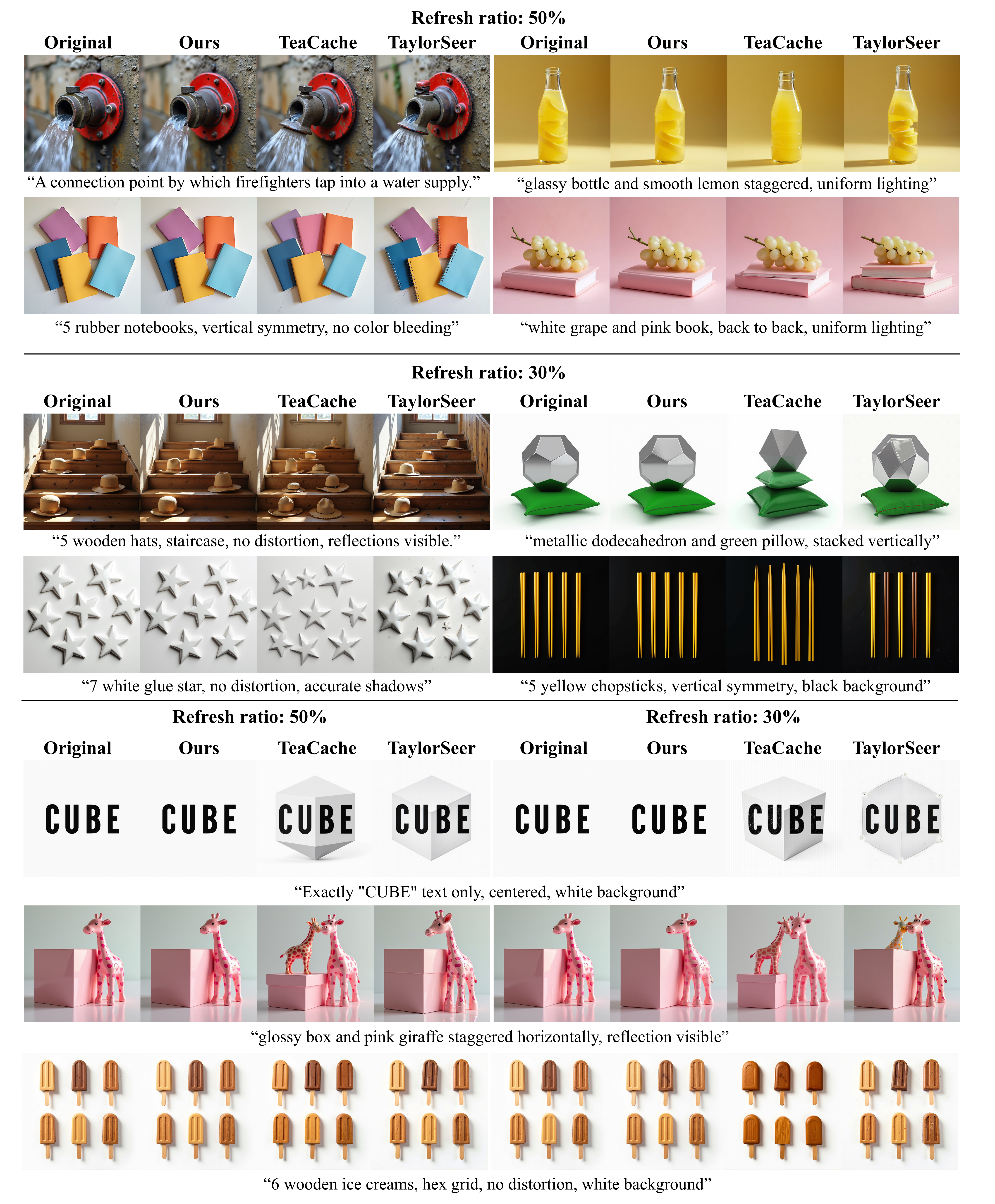}
  \vspace{-10pt}
  \caption{
  Additional qualitative comparison of SeaCache and baselines on \textit{FLUX} at refresh ratios of approximately \(30\%\) and \(50\%\).
  }
  \label{fig:supp_qual_image}
\end{figure*}

In Fig.~\ref{fig:supp_qual_image}, we provide additional qualitative comparisons on \textit{FLUX} at refresh ratios of approximately 50\% (top panel) and 30\% (middle panel), along with an additional set of examples at both cache budgets in the bottom panel.

At 50\% refresh ratio in the top-left of Fig~\ref{fig:supp_qual_image}, SeaCache preserves a clean water surface without the blocky artifacts or texture distortions that appear in the baselines.
In the top-right example, the baselines either generate a blurry lemon or fail to capture the fluid dynamics inside the bottle, whereas SeaCache correctly synthesizes both the glass bottle and the orange liquid, closely matching the full-compute original reference.

At a more aggressive 30\% refresh ratio in the middle panel, SeaCache again stays closest to the full-compute reference. In the middle-left example, only SeaCache reconstructs seven well-formed stars consistent with the original, while competing methods either miss or severely deform several stars.
In the middle-right example, SeaCache produces five chopsticks with consistent length and color, whereas the baselines generate chopsticks with mismatched geometry and appearance.

In the bottom panel of Fig.~\ref{fig:supp_qual_image}, we further compare the same text prompts across different cache budgets using the same seed. In the top row of the panel, for the prompt requesting exactly the word ``CUBE,'' the baselines repeatedly hallucinate cube-like patterns in the background, whereas SeaCache is the only method that successfully renders the intended text. In the last row of the panel, all methods generate six wooden ice creams, but the baselines produce slightly different designs or colors compared to the full-compute reference, while SeaCache most closely matches the original design.

These additional cases further support that SeaCache best preserves the original content and layout while operating under the same cache budgets.

\subsection{Qualitative Comparison in T2V Generation}

\begin{figure*}[t!]
    \vspace{-2pt}
  \centering
  \includegraphics[width=0.96\textwidth]{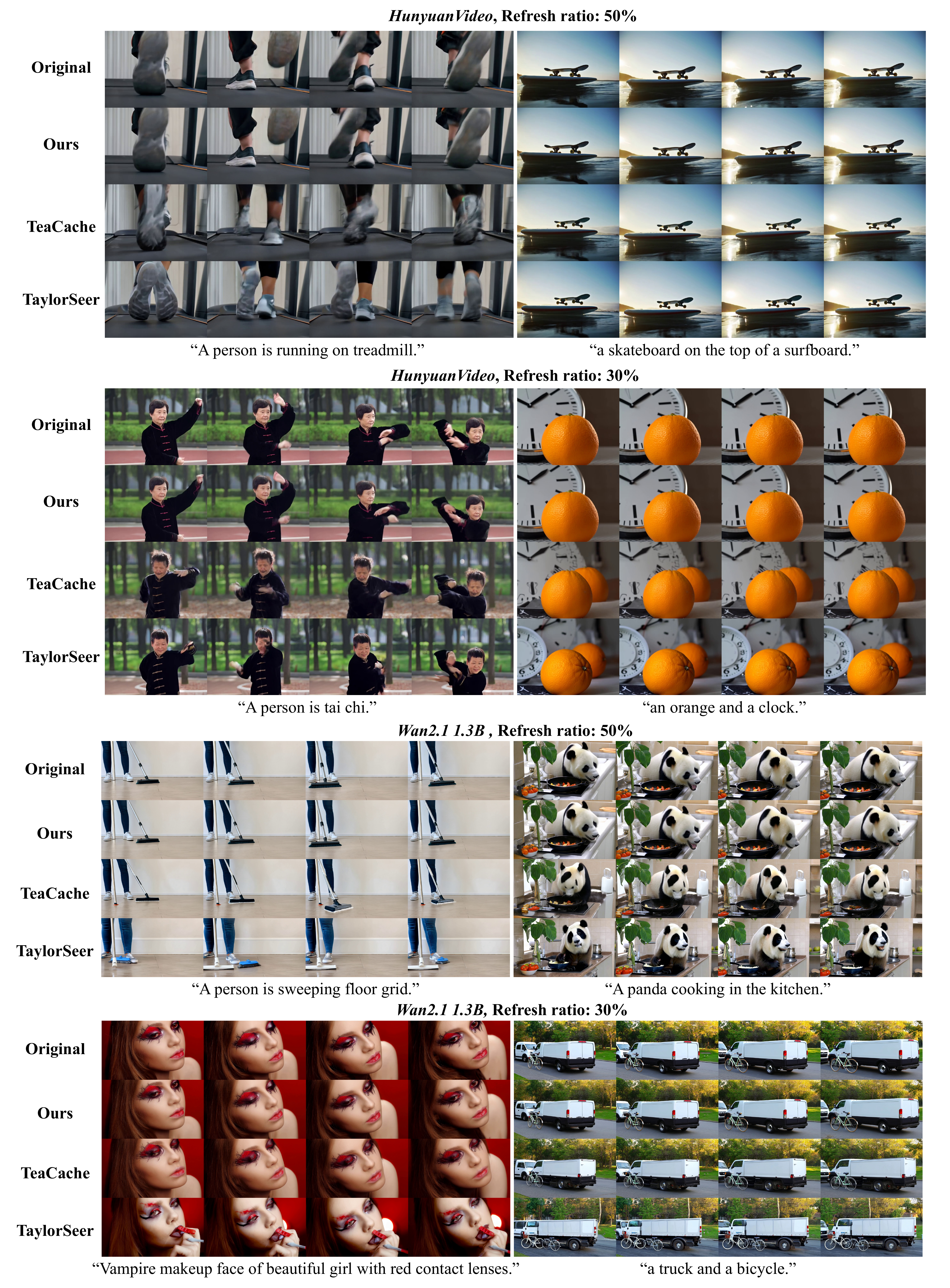}
  \vspace{-10pt}
  \caption{
  Additional T2V qualitative comparison of SeaCache and baselines at refresh ratios of approximately \(30\%\) and \(50\%\).
  }
  \label{fig:supp_qual}
\end{figure*}

Fig.~\ref{fig:supp_qual} presents further qualitative comparisons on \textit{HunyuanVideo} and \textit{Wan2.1 1.3B}, respectively.
For each prompt, we horizontally concatenate the same intermediate frame index from the full-compute reference and all caching variants to isolate per-frame differences.
On \textit{HunyuanVideo} at a \(30\%\) refresh ratio, the baselines exhibit severe artifacts around the hands during the Taichi motion, while SeaCache preserves a plausible pose with smooth limb contours.
At \(50\%\) refresh, the baselines render a skateboard that appears to float above the surfboard, whereas SeaCache correctly places the skateboard in contact with the surfboard, matching the original video, as shown in the right side of Fig.~\ref{fig:supp_qual}.

On \textit{Wan2.1 1.3B} at a \(30\%\) refresh ratio the baselines introduce noticeable distortions near the truck wheels and bicycles, but these artifacts do not appear in the SeaCache outputs, as visualized in Fig.~\ref{fig:supp_qual}.
At \(50\%\) refresh, competing methods either cause food items on the table to disappear or introduce artifacts on the panda, while SeaCache closely follows the full-compute trajectory without these failures.
Overall, these qualitative results indicate that SeaCache better tracks the original dynamics and adheres more faithfully to the text prompts while avoiding objectionable artifacts.


\section{Limitation}
\label{sec:limitation}
To derive the optimal linear filter, we adopt several simplifying assumptions that make the spectral response analytically tractable, even though they need not hold exactly in practice. We model the signal spectrum with a power law under a radial view, whereas generated samples, particularly at later timesteps or in highly synthetic backgrounds with no salient objects, can deviate from this behavior. We also assume wide-sense stationarity and independence between signal and noise. When these conditions are violated, the closed-form linear filter is no longer strictly optimal and can introduce bias.

In addition, our analysis is formulated in the image or video domain, while most modern generative models operate in a learned latent space. The encoder can reshape the spectrum, so the latent distribution may differ from the assumed pixel-domain power-law model, and our filter then only approximates the optimal latent-space response.

A promising extension is to relax these assumptions by estimating per-timestep spectra, designing content-aware filters directly in the latent space, and augmenting them with lightweight nonlinear corrections, while preserving the plug-and-play nature of our cache policy. These extensions would reduce the gap between the assumed and actual signal models and further improve fidelity under real-world deviations from our assumptions.

\end{document}